\pdfoutput=1
\documentclass[11pt]{article}

\usepackage[final]{acl}

\usepackage{times}
\usepackage{latexsym}
\usepackage{booktabs}
\usepackage{float}
\usepackage{stfloats}
\usepackage{listings}
\usepackage{enumitem}

\usepackage[T1]{fontenc}
\usepackage[utf8]{inputenc} 

\usepackage{microtype}

\usepackage{inconsolata}
\usepackage{custom_commands}
\usepackage{tabularx}
\usepackage{cleveref}

\usepackage{amsmath,amsfonts,bm}

\def\eqref#1{equation~\ref{#1}}
\def\1{\bm{1}}

\def\vx{{\bm{x}}}

\def\vz{{\bm{z}}}

\DeclareMathAlphabet{\mathsfit}{\encodingdefault}{\sfdefault}{m}{sl}
\SetMathAlphabet{\mathsfit}{bold}{\encodingdefault}{\sfdefault}{bx}{n}

\newcommand{\E}{\mathbb{E}}

\newcommand{\KL}{D_{\mathrm{KL}}}
\newcommand{\Var}{\mathrm{Var}}

\title{Exploring Precision and Recall\\ to assess the quality and diversity of LLMs}

\author{Florian Le Bronnec$^{*,1,2}$ \quad  {\bf Alexandre Verine$^{*,1}$} \\ {\bf Benjamin Negrevergne$^1$} \quad {\bf Yann Chevaleyre$^1$} \quad{\bf Alexandre Allauzen$^1$}\\
\small$^1$Miles, Université Paris-Dauphine, Université PSL, CNRS, LAMSADE, 75016 Paris, France \\
\small$^2$Sorbonne Université, CNRS, ISIR, F-75005 Paris, France}
\begin{document}
\maketitle
\def\thefootnote{*}\footnotetext{Authors contributed equally to this work. Corresponding authors: \texttt{florian.le-bronnec@dauphine.psl.eu, alexandre.verine@dauphine.psl.eu}.}
\renewcommand{\thefootnote}{\arabic{footnote}}
\begin{abstract}
    We introduce a novel evaluation framework for Large Language Models (LLMs) such as \textsc{Llama-2} and \textsc{Mistral}, focusing on importing Precision and Recall metrics from image generation to text generation. This approach allows for a nuanced assessment of the quality and diversity of generated text without the need for aligned corpora. By conducting a comprehensive evaluation of state-of-the-art language models, the study reveals new insights into their performance on open-ended generation tasks, which are not adequately captured by traditional benchmarks. The findings highlight a trade-off between the quality and diversity of generated samples, particularly when models are fine-tuned on instruction dataset or with human feedback. This work extends the toolkit for distribution-based NLP evaluation, offering insights into the practical capabilities and challenges that current LLMs face in generating diverse and high-quality text. We release our code and data\footnote{\url{https://github.com/AlexVerine/pr-4-llm}}.
\end{abstract}

\section{Introduction}

%  In the recent years Large Language Models (LLMs) have been popularized by chat interfaces such as ChatGPT or Hugging-Chat, and are now used for a variety of tasks including writing emails,  application letters, and even writing novels or poems. 

In recent years and months, there has been a rapid democratization of Large Language Models (LLMs), exemplified by platforms such as ChatGPT \citep{openai2024gpt4} and HuggingChat \citep{huggingface_chatui}. These models are now widely accessible for a diverse range of tasks, from composing emails and application letters to performing multi-document summarization, generating medical prescriptions, and even crafting poetry and novels \citep{hosseini2023exploratory,skjuve2023user}. The expanding spectrum of applications underscores the ubiquity of LLMs in our daily lives, necessitating the development of novel evaluation frameworks to accommodate their growing relevance.

\begin{figure}[ht!]
    \centering
    \includegraphics[width=\linewidth]{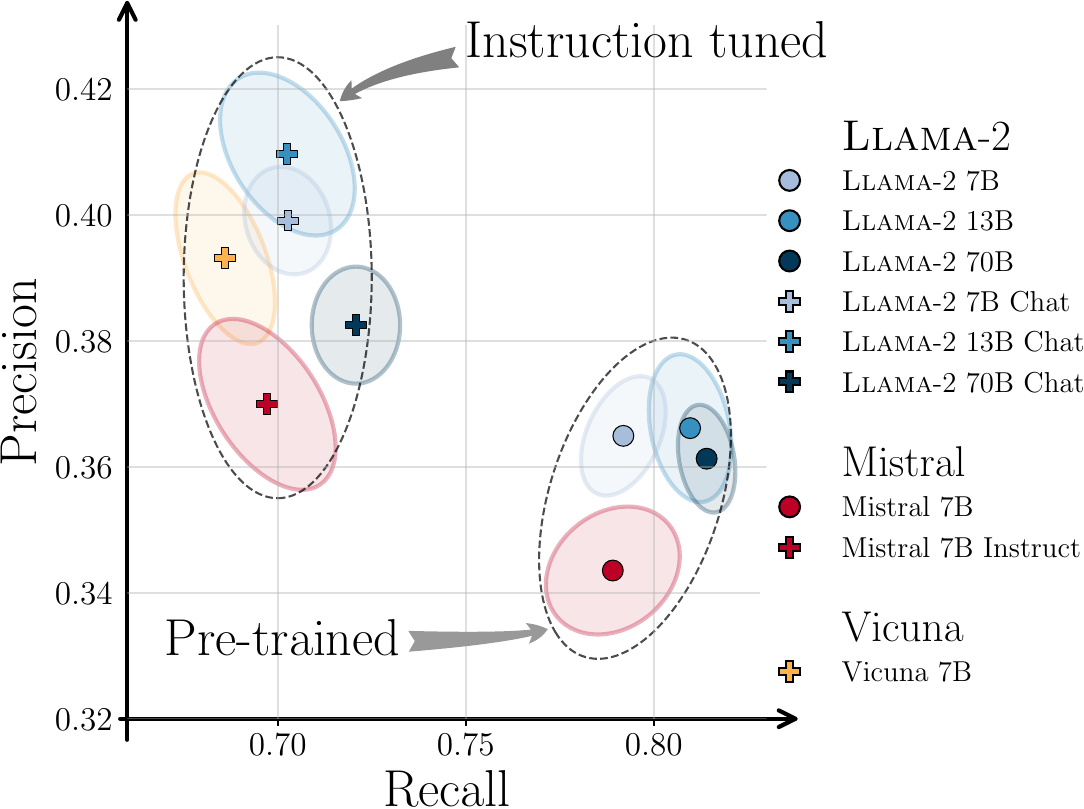}
    \caption{Precision and Recall of various models on generating the WebText dataset, with the 2 standard deviation error ellipsis. Chat and pre-trained models different behaviors are clearly captured by oevidenced.}
    \label{fig:webtext-PR}
\end{figure}
Until recently, benchmarks were designed to target specific tasks such as machine translation, summarization, and question answering, among others. However, LLMs now encompass a wide range of tasks, prompting the community to reconsider methods for comparing and assessing these models. One proposed solution is to gather a diverse set of tasks to better evaluate the versatility of modern generative models. For example, the Open LLM Leaderboard \citep{open-llm-leaderboard} presents a unified framework consisting of closed questions that focus on the model's ability to provide concise and accurate answers. However, it is important to note that these evaluations are sample-based, relying on generated samples from the model compared against aligned references, written by humans.

In contrast, the recent line of work on \emph{distribution-based} metrics greatly departs from the \emph{sampled based} evaluation.  By considering LLMs and datasets as empirical distributions, the new metrics attempt to quantify how they differ and how they overlap. This shift drastically changes the scope of the evaluation. Beyond performance measures based on human references, the objective is to estimate the disparity between some data distribution ($P$) exhibited by human-written texts and the distribution learned by a LLM ($Q$), eliminating the need for
aligned corpora. As exemplified by the development of MAUVE \cite{pillutla_mauve_2021}, this kind of approach opens new perspectives for comparing LLMs in terms of their generative abilities rather than for some peculiar tasks.

% In contrast, a recent shift in focus among researchers has led to the exploration of distribution-based metrics, exemplified by the development of MAUVE \cite{pillutla_mauve_2021}. Unlike traditional evaluation techniques, distribution-based methods aim to quantify the divergence between the actual data distribution ($P$) and the model's distribution ($Q$), eliminating the need for aligned corpora. Building upon this novel approach, our study extends the concepts of Precision and Recall, which have previously been adapted to distributions in domains such as image generation by \cite{} and \cite{}. We demonstrate that Precision and Recall offer effective measures for capturing two prevalent types of modeling errors in LLMs: those resulting in low-quality samples and those associated with a lack of diversity in generated outputs.

However, many factors take part in assessing open-ended text generation and focusing on a single measure may restrict the significance of the evaluation. This is well known in the field of Information Retrieval, for example, where Precision and Recall are at the core of evaluation.
Although the usage of Precision and Recall for generative models has been previously used in image generation by \citet{sajjadi_assessing_2018} and \citet{kynkaanniemi_improved_2019}, we propose to adapt them to LLMs. With an extensive set of experiments involving prominent LLMs such as \textsc{Llama-2} \citep{touvron2023llama} and Mistral \citep{mistral}, we show that these two new metrics significantly improve the evaluation with a better understanding and characterization of the flaws in text generation. Precision and Recall allow us to clearly distinguish between samples quality or adequacy (for Precision) and a lack of diversity in the model outputs (for Recall). Empirical results
show that these two measures are necessary for a in-depth comparison of LLMs.  For instance, we are able to quantify that fine-tuning models on instruction sets with human feedback significantly improves sample quality, albeit at the expense of sample diversity, as evidenced in \Cref{fig:webtext-PR} by the trade-off between Precision and Recall.

%In this paper, we leverage this innovative evaluation framework to assess the performance of prominent language models like Llama and Mistral, using Precision and Recall as our metrics. Through our analysis, we uncover several compelling insights into the capabilities of these models. For instance, our experiments reveal that fine-tuning models with human feedback can significantly enhance sample quality, albeit at the expense of sample diversity, as evidenced by the trade-off between Precision and Recall.
As a summary, our contributions are threefold:

\begin{itemize}[leftmargin=*] \setlength{\itemsep}{0em}
    \item We adapt the concepts of Precision and Recall, traditionally used in image generation, to evaluate the performance of large language models. Our method offers a novel lens to assess the quality and diversity of text generation without the need for aligned corpora.%through which the quality and diversity of text generation can be assessed without the need for aligned corpora.
    \item We carry out a thorough evaluation of state-of-the-art language models, such as \textsc{Llama} and \textsc{Mistral}, using our proposed framework. This analysis provides a detailed comparison of the performance of these models in terms of quality and diversity on challenging open-ended generation tasks that are out of the scope of traditional evaluation methods.
    \item  Our investigation sheds light on the impact of fine-tuning LLMs as chat models. We present empirical evidence that, while such fine-tuning can improve the quality of generated samples, it also reduce their diversity, highlighting a crucial trade-off between Precision and Recall.
\end{itemize}

In summary, our contributions advance the field of distribution-based NLP evaluation by introducing novel metrics tailored to the specific tasks of open-ended generation. Through empirical analysis and insights, we provide a deeper understanding of LLM generation capabilities.

\begin{figure*}[t]
    \centering
    \subfloat[t][Distributions of labels.]{\includegraphics[width=0.25\textwidth]{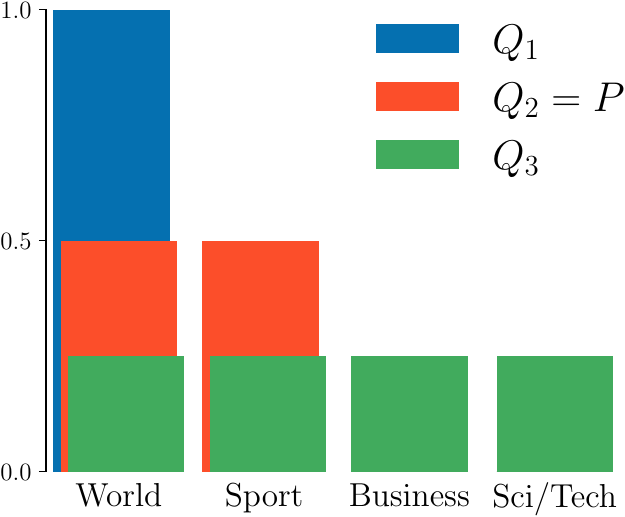}\label{fig:agnews_labels}}\hspace{0.03 \linewidth}
    \subfloat[t][MAUVE of $Q$ w.r.t.\ $P$.]{\includegraphics[width=0.21\textwidth]{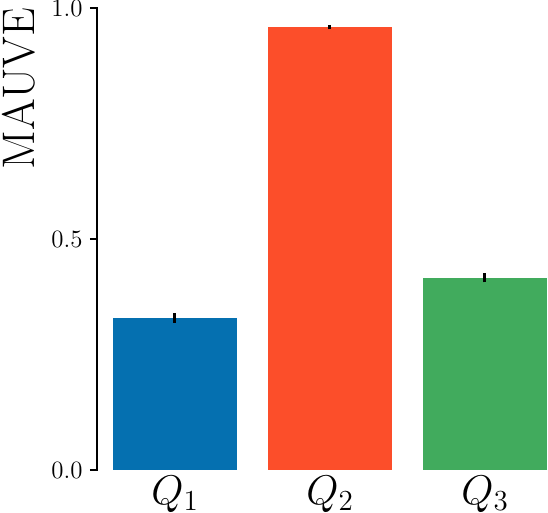}\label{fig:agnews_mauve}}\hspace{0.03 \linewidth}
    \subfloat[t][Precision of $Q$ w.r.t.\ $P$.]{\includegraphics[width=0.21\textwidth]{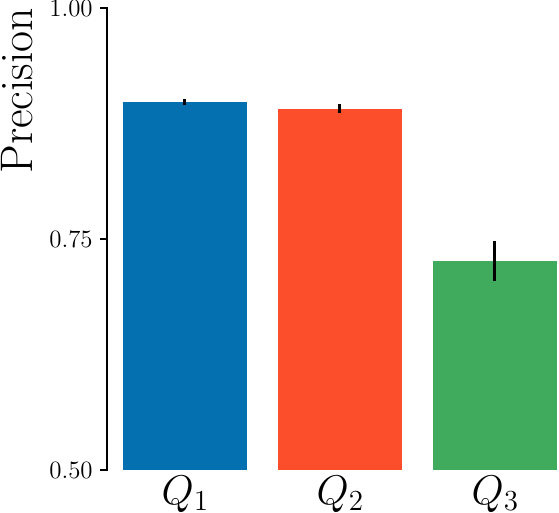}\label{fig:agnews_precision}}\hspace{0.03 \linewidth}
    \subfloat[t][Recall of $Q$ w.r.t.\ $P$.]{\includegraphics[width=0.21\textwidth]{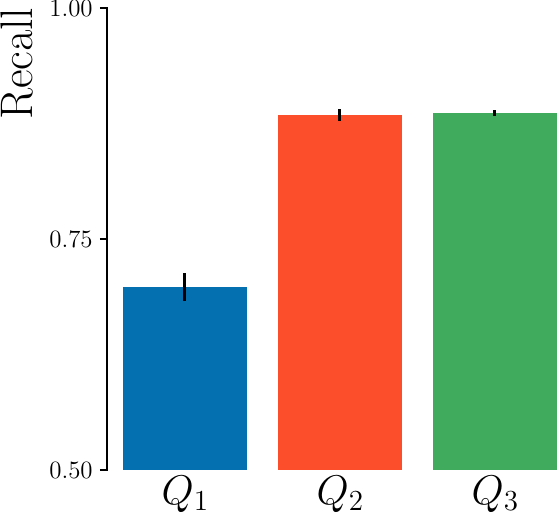}\label{fig:agnews_recall}}
    \caption{Illustration of a simple case where using MAUVE score alone fails to provide a fine-grained evaluation of quality and diversity. We consider a reference dataset $P$ composed of articles from 2 labels, \texttt{World} and \texttt{Sport}. $Q_2$ is made of articles from the same distribution. We compare it with two other datasets: $Q_1$ composed only of \texttt{World} articles and $Q_3$ composed of even numbers of \texttt{World, Sport, Business} and \texttt{Sci/Tech} articles. Relatively to $Q_2$, the MAUVE scores of $Q_1$ and $Q_3$ are almost identical, while Precision and Recall help differentiating how the distributions actually differ from the reference $P$.}
    \label{fig:agnews}
\end{figure*}

%The distribution of labels shown in \ref{fig:agnews_labels} is used to build $P$, $Q_1$, $Q_2$ and $Q_3$. The MAUVE score between $P$ and $Q$ is shown in \ref{fig:agnews_mauve}, the Precision in \ref{fig:agnews_precision} and the Recall in \ref{fig:agnews_recall}. Metrics are computed on 5000 random samples over 5 seed and standard deviation is showed in black. MAUVE is not able to differentiate the dataset with 50\% of the required labels and the dataset with 50\% of undesired labels.  

\section{Related works}

Historically, assessing text generation tasks has been dependent on human resources, such as annotated texts or aligned corpora. This is particularly true in areas like machine translation and automatic summarization, where N-gram based metrics have been developed, notably BLEU \citep{papineni_bleu_2001} and ROUGE \citep{lin_rouge_2004}. These metrics perform a rough comparison between the generated text and human-produced and aligned reference texts. More recent developments have introduced model-based metrics, such as BERTScore\footnote{BERTScore uses precision and recall scores that differ significantly from ours, as they evaluate the similarity between a single generated text and its aligned reference.} \citep{bertscore}, which aim to capture semantic similarities more effectively. However, these evaluations still largely depend on "aligned references," meaning human-provided examples that constrain the expected range of generated outputs. The challenge becomes significantly greater with open-ended generation tasks, where the range of acceptable responses is vast and cannot be encapsulated by a single reference.
\subsection{Standard evaluation of generative models}
%\paragraph{Assessment of Large Language Models (LLMs).}

Perplexity is the historical measure for language modeling
\cite{Bahl77}. Easy to use and cheap to compute, practical motivations clearly explain its persistence for evaluation, despite its notable
limitations. As demonstrated in the work of \citet{pillutla_mauve_2021},
perplexity works at the token level, only considering the surface
forms, hence missing semantic information.  Moreover, the metric is
calculated on a per-sample basis, thereby failing to capture the
diversity of generated texts.

Nowadays, the prevalent method for evaluating recent LLMs involves
benchmarks with many tasks with easily verifiable correct
answers
\citep{touvron2023llama,geminiteam2023gemini,open-llm-leaderboard}. Typically,
this involves the generation of responses to short questions prompted
in the instructions. The responses are then compared to a reference
using metrics such as Exact-Match. While effective in assessing the
comprehension and reasoning capabilities of LLMs, this kind of
approach does not assess the generation skills of the models.

To overcome the limitations of automatic metrics, another trend considers human evaluation \citep{celikyilmaz2021evaluation}, especially for complex tasks.  Beyond its cost, the guidelines must be carefully designed, since various criteria could be necessary to rate texts and compare model outputs. In the end, human evaluation is prone to high variance and is difficult to reproduce. Another issue is that diversity is not evaluated is standard human evaluation protocols, since texts are rated individually or by pairs.

\subsection{Evaluation of open-ended generation}

%\paragraph{Diversity oriented metrics.}

Although the prevailing evaluations focus primarily on the quality of generated texts, some studies aim to characterize the diversity of these outputs. An approach involves computing the proportion of distinct N-grams generated, as demonstrated in \cite{distinctn}. Another notable proposition by \cite{zhu2018texygen} introduces Self-BLEU which computes the BLEU score of each generated text against all other generated texts as a reference. However, N-gram matching cannot capture semantic features. For instance, a model generating random words achieves a perfect Self-BLEU score.

\subsection{Quality/diversity tradeoff}
The tradeoff between quality and diversity has been studied by \citet{go2023aligning} and \citet{kirk2024understanding} specifically during the human feedback alignment phase of LLMs. They draw their conclusions through traditional metrics and specifically through the summarization task. By introducing new metrics for such evaluation and focusing on extensive open-ended tasks, our methods bring new tools inspired by past research in image generation and backed by a solid theoretical framework. Our findings extend throughout the human feedback alignment phase to include both instruction tuned and pre-trained, highlighted significant differences between those types of models.

% \paragraph{Distribution-Based Metrics.}

In the current landscape, most of the existing methods prioritize quality or rely on a simplistic characterization of diversity. Nevertheless, the recent introduction of MAUVE \citep{pillutla_mauve_2021} makes a significant shift toward diversity. Drawing inspiration from the domain of image generation, this new metric compares generated texts to a reference distribution without requiring aligned texts. MAUVE consists of a divergence-based metric that captures certain properties of text and exhibits correlation with human judgment. Subsequent research by \citet{pimentel2023usefulness} confirms the validity of this approach by leveraging common divergence measures. MAUVE and subsequent works propose a simple metric that allows us to summarize the evaluation of both quality and diversity in a single measure. Our work builds upon these efforts; we aim to demonstrate that using two distinct measures enables us to distinguish between lack of quality and lack of diversity. This distinction becomes necessary to enable a deeper and more precise understanding of generative models, as illustrated by the simple example of \Cref{fig:agnews}.

\begin{figure}[t!]
    \centering
    \subfloat[$Q_1$. P $\approx 0.95$, R $\approx 0$.]{\includegraphics[width=0.46\linewidth]{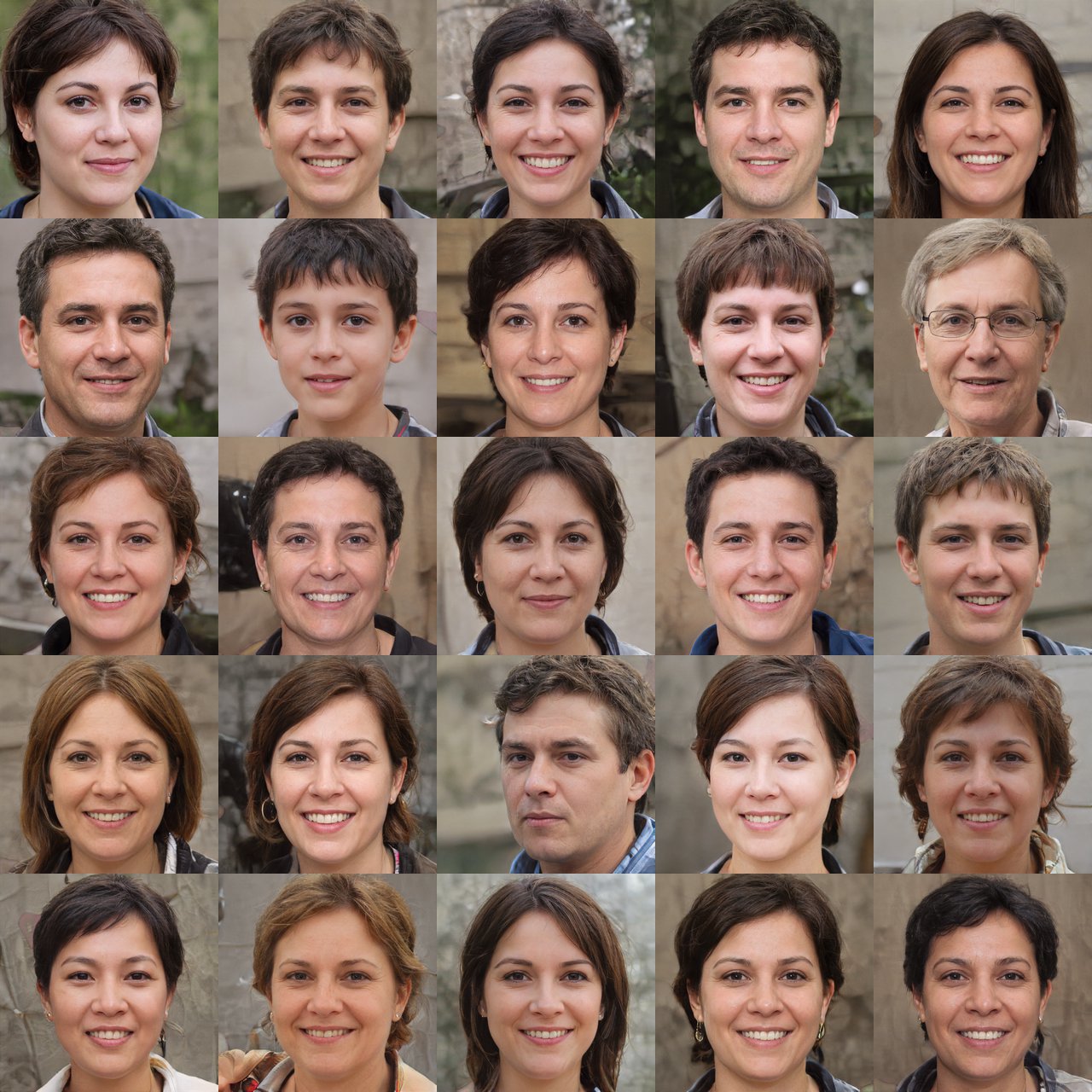}}\hspace{0.05\linewidth}
    \subfloat[$Q_2$. P $\approx0.4$, R $\approx 0.4$.]{\includegraphics[width=0.46\linewidth]{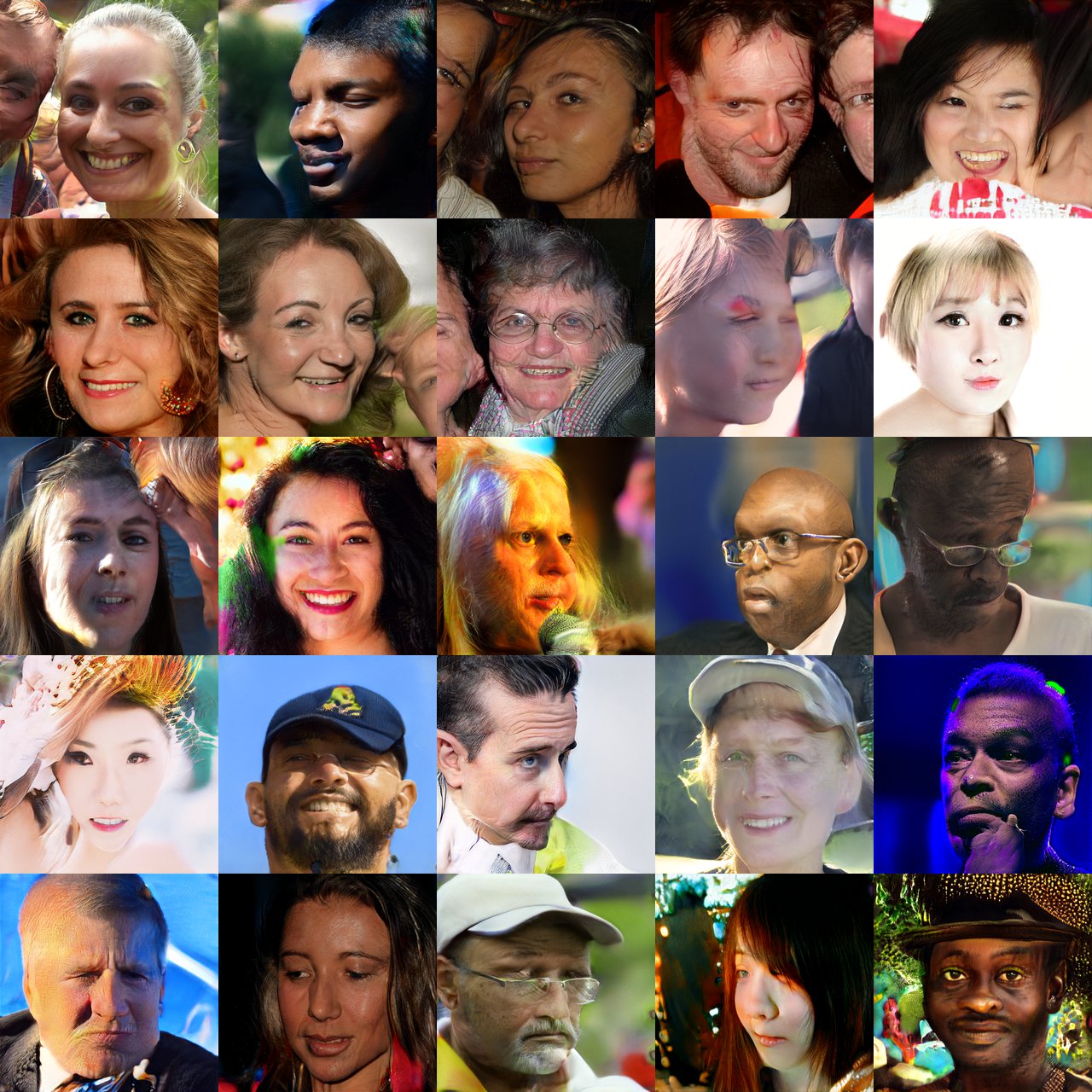}}
    \caption{Example of distribution of images. $P$ is the reference distribution of images of the CelebA dataset \cite{liu_deep_2015}, $Q_1$ and $Q_2$ are two different distributions of images. $Q_1$ has high quality, but low diversity, while $Q_2$ has high diversity and low quality. Numbers and images are from \citet{kynkaanniemi_improved_2019}.}
    \label{fig:celeba}
\end{figure}

\section{Background on Precision and Recall}
\label{sec:prdef}
\begin{figure*}[hb!]
    \centering
    \subfloat[Distributions $P$ and $Q$]{\includegraphics[width=0.27\linewidth]{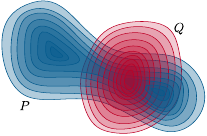}}\hspace{0.03\linewidth}
    \subfloat[Precision to assess quality]{\includegraphics[width=0.27\linewidth]{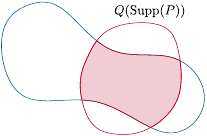}}\hspace{0.03\linewidth}
    \subfloat[Recall to assess diversity]{\includegraphics[width=0.27\linewidth]{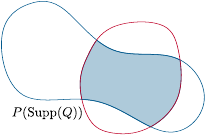}}
    \caption{Precision and Recall for distribution-based metrics. (a) Distributions $P$ and $Q$. (b) Precision is the proportion of the support of $Q$ that generates $P$. (c) Recall is the proportion of the support $P$ generated by $Q$.}
    \label{fig:pr_distributions}
\end{figure*}

%In this section, we introduce the theoretical background behind the Precision and Recall for distribution dissimilarity measures, and we present how they are used to assess quality and diversity of generative models.
%\paragraph{Introduction of Precision and Recall.}
Nowadays generative models for images are capable of producing ultra-realistic samples, making classical metrics such as Fr\'echet Inception Distance (FID) \citep{heusel_gans_2017} too coarse to assess the full
distributional properties of the models. This jump of performance has motivated the definition of precision and diversity for generative models.
This need for a finer evaluation is exemplified on \Cref{fig:celeba}. $Q_1$ represents a model that produces realistic images but lacks diversity in the generated people's appearance. Conversely, $Q_2$ depicts a model generating a broad range of people with lower quality. Capturing this significant tradeoff is essential for comprehending generative model behavior.
Inspired by the Precision and Recall metrics of binary classifications,
the pioneering work of \citet{sajjadi_assessing_2018} proposed to redefine these two terms  to assess the quality and diversity of generative models. These metrics are  based on a discretized estimation of the reference and model's distribution. \citet{kynkaanniemi_improved_2019} pointed out some limitations of those metrics and proposed a new way to compute Precision and Recall, leveraging more robust support estimation. This method is now widely adopted in the literature and is typically used along with the FID. Still, these metrics have some limitations, like the sensibility to outliers \citep{naeem_reliable_2020}. This motivates further ongoing work on this topic, including theoretical contributions \citep{simon_revisiting_2019,naeem_reliable_2020,alaa_how_2022,cheema_precision_2023, verine_precision-recall_2023} or practical methods \citep{kim_toppr_2023}.

\paragraph{Precision and Recall definition.} Given its importance in the field of image generation, we adopt the definition of Precision and Recall introduced by \citet{kynkaanniemi_improved_2019}. Alternative definitions and approaches are discussed in Appendix~\ref{sec:prcurvedef}.
\begin{definition}
    Let $P$ and $Q$ be two distributions over a space $\mathcal{X}\in\mathbb{R}^d$. We denote $\Supp(P)$ and $\Supp(Q)$ their support. The Precision and the Recall of $Q$ with respect to $P$ are defined as:
    \begin{subequations}
        \begin{align}
            \mathrm{Precision} & = Q(\Supp(P))  \\
            \mathrm{Recall}    & = P(\Supp(Q)).
        \end{align}
    \end{subequations}
\end{definition}

\paragraph{Precision and Recall in practice.} Since Precision and Recall involve an estimation of the support, standard pipelines involve a projection of the samples into an latent space, using pre-trained models such as Inception-v3 \citep{szegedy_rethinking_2015} or VGG \citep{simonyan_very_2015}. Supports of the distributions are then estimated using a $k$-nearest neighbors algorithm. We present this algorithm in more detail in \Cref{sec:pr-text}.

\paragraph{Quality and diversity in NLP.}
Although using metrics characterizing quality and diversity at a distribution level is well-established in image generation \citep{dhariwal_diffusion_2021, sauer_stylegan-xl_2022, song_consistency_2023}, it is a very recent practice in the field of text generation. \citet{pillutla_mauve_2021} led the way in this field by introducing MAUVE, a pioneering metric that builds on the theoretical concepts of quality and diversity, as initially proposed by \citet{djolonga_precision-recall_2020}, and detailed in Appendix~\ref{sec:prcurvedef}.

The authors showcased the intriguing characteristics of their metrics across various GPT-2 models and decoding algorithms. This metric is highly effective in evaluating the degree to which the generated distribution aligns with the reference. The methodology behind it was subsequently validated and simplified by \citet{pimentel2023usefulness} and further evaluated with alternative frameworks by \citet{pillutla_mauve_2023}.

Despite the progress made, these approaches present a challenge: interpreting them as a measure of quality and diversity is not straightforward, as they condense these two concepts into a single divergence measure.  Characterizing Precision and Recall independently has only been mentioned by \citet{pillutla_mauve_2023} who conducted some toys experiments on small models or decoding algorithms.

Therefore, we believe that Precision and Recall for text generation are in its early stages. We argue that a deeper understanding of these metrics and their practical implications is needed.

\section{Precision and Recall to assess text generation}
\label{sec:pr-text}
In this paper, we introduce a framework for calculating Precision and Recall between distributions on texts. These distributions characterize a dataset, or a model, and its generative capability. The extended version of Precision and Recall builds upon the notion of support of distributions, which is in our case challenging to define and estimate. 
% text datasets, with the primary aim of applying these metrics to generated text. We begin by outlining the general pipeline  and then showcase how we determined the specific parameters.

\begin{figure*}[ht!]
    \centering
    \includegraphics[width=\textwidth]{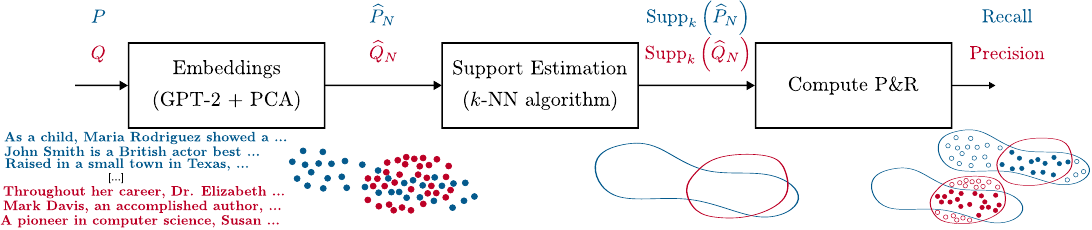}
    \caption{Our pipeline to compute the Precision and Recall metrics. Texts are projected into a latent space of a  pre-trained model, where a $k$-NN estimation is performed to estimate the relative overlaps of $P$ and $Q$.}
    \label{fig:pipeline}
\end{figure*}

\subsection{Pipeline to compute Precision and Recall}
We propose the following pipeline to compute the Precision and Recall metrics, as shown in Figure~\ref{fig:pipeline}:
\begin{enumerate}[leftmargin=*]\setlength{\itemsep}{0em}
    \item \textbf{Sampling.} Select $N$ sequences randomly in the dataset to build the set of references: $\mathcal{X}^\mathrm{ref}=\left\{\vx_1^{\mathrm{ref}}, \dots, \vx_N^{\mathrm{ref}}\right\}.$ Sample $N$ sequences from the distribution $Q$ to assess and build the set of outputs: $\mathcal{X}^\mathrm{out}=\left\{\vx_1^{\mathrm{out}}, \dots, \vx_N^{\mathrm{out}}\right\}.$
    \item \textbf{Samples pre-processing.}
          \begin{enumerate}[leftmargin=*]
              \item Compute the latent representations of each set using the embedding function $\phi$:
                    \begin{align*}
                        \mathcal{X}^\mathrm{ref}_{\phi} & =\left\{\phi(\vx_1^{\mathrm{ref}}), \dots, \phi(\vx_N^{\mathrm{ref}})\right\} \\
                        \mathcal{X}^\mathrm{out}_{\phi} & =\left\{\phi(\vx_1^{\mathrm{out}}), \dots, \phi(\vx_N^{\mathrm{out}})\right\}
                    \end{align*}
              \item Perform Principal Component Analysis (PCA) on the union of sets $\mathcal{X}^\mathrm{ref}_{\phi}$ and $\mathcal{X}^\mathrm{out}_{\phi}$ to reduce the dimensionality of the data by retaining 90\% of the variance. This process yields the sets $\mathcal{Z}^\mathrm{ref}_\phi$ and $\mathcal{Z}^\mathrm{out}_\phi$, which consist of points, respectively, distributed on $\widehat{P}_N$ and $\widehat{Q}_N$, the empirical distributions in the latent space.
          \end{enumerate}
    \item \textbf{Support estimation.}
    
          \begin{enumerate}
              \item For each set $\mathcal{Z}_\phi\in\left\{\mathcal{Z}^\mathrm{ref}_\phi,\mathcal{Z}^\mathrm{out}_\phi\right\}$, for each point $\vz_i \in \mathcal{Z}_\phi$ computes the pairwise distances. Define $B_k(\vz_i,\mathcal{Z}_\phi)$ as the ball centered on $\vz_i$ with radius being the distance to the $k$-th nearest neighbor in $\mathcal{Z}_\phi$.
              \item Each support is defined as the union of balls $B_k(\vz_i,\mathcal{Z}_\phi)$ for all $\vz_i$ in $\mathcal{Z}_\phi$:
                    \begin{align*}
                        \Supp_k(\widehat{P}_N) & = \bigcup\nolimits_{\vz_i \in \mathcal{Z}^\mathrm{ref}_\phi} B_k(\vz_i,\mathcal{Z}^\mathrm{ref}_\phi) \\
                        \Supp_k(\widehat{Q}_N) & = \bigcup\nolimits_{\vz_i \in \mathcal{Z}^\mathrm{out}_\phi} B_k(\vz_i,\mathcal{Z}^\mathrm{out}_\phi)
                    \end{align*}
          \end{enumerate}
    \item \textbf{Precision and Recall computation.}
          \begin{subequations}
              \begin{align}
                  \mathrm{Precision} & = \frac{1}{N} \sum_{\vz_i \in \mathcal{Z}^\mathrm{out}_\phi} \mathds{1}_{\vz_i \in \Supp_k(\widehat{P}_N)} \\
                  \mathrm{Recall}    & = \frac{1}{N} \sum_{\vz_i \in \mathcal{Z}^\mathrm{ref}_\phi} \mathds{1}_{\vz_i \in \Supp_k(\widehat{Q}_N)}
              \end{align}
          \end{subequations}
\end{enumerate}
The parameters are detailed in the next section.
\begin{figure}[b!]
    \centering
    \includegraphics[width=0.42\textwidth]{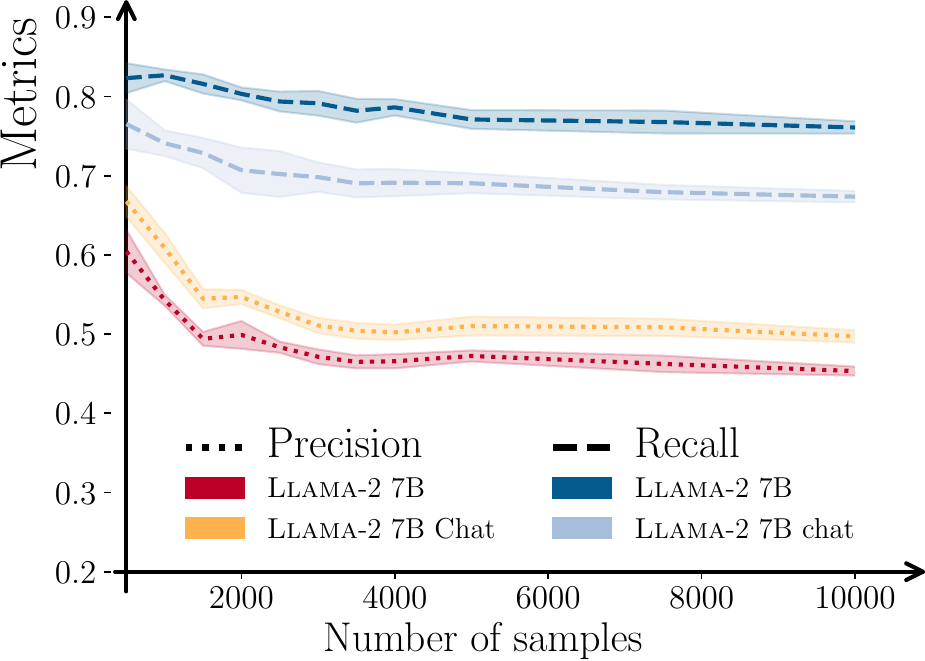}
    \caption{Evolution of the metrics Precision and Recall computed with $k=4$ as the number of samples $N$ increases. The shaded area represents the standard deviation computed over five random seeds for the set outputs. The means reach a plateau around $N=3000$.}
    \label{fig:diversity_numsamples}
\end{figure}
\subsection{Precision and Recall for text generation.}

The embeddings used for the MAUVE metric have proven to be effective at capturing the properties of the text at the word level \citep{pimentel2023usefulness} and content level \citep{pillutla_mauve_2021}. Building on this success, we adopt the same embedding function, $\phi$, which leverages the output of the last layer of $\text{GPT-2}_{\text{LARGE}}$, coupled with a PCA. The PCA serves a dual purpose: it reduces data dimensionality and also filters out noise, which is particularly beneficial for $k$-NN based manifold estimation. As presented in \Cref{app:other-embeddings}, the use of other models for the embeddings gives consistent results.

To choose the other parameters -- the number of samples $N$ and the neighbor parameter $k$ -- we investigate how Precision and Recall are affected by their variations. We use the WebText generation task with \llamaSchat (comprehensively described in Section~\ref{sec:use-cases:open-ended}) as a test case, prompting the model to generate between 100 and 10,000 samples. The findings are depicted in \Cref{fig:diversity_numsamples}. Additional details can be found in \Cref{app:webtext}.

Our observations are as follows: (1) Precision and Recall values stabilize with an increasing $N$, plateauing at approximately $N=3000$. (2) As $k$ increases, Precision and Recall values gradually approach 1, aligning with the expectation that a larger $k$ leads to a wider estimated support.
In the rest of the paper, we set $N$ to $4000$ samples and $k$ to $4$.

\section{Practical use cases for LLMs}
Equiped with our freshly introduced Precision and Recall we now present an analysis of the quality of generation of a broad range of LLMs. We focus on three different tasks to illustrate the versatility of our metrics and the new insights they can provide.

% On three open-ended generation tasks, 1) we provide evidence that our metrics can capture the trade-off between the quality and diversity of generated samples, 2) we use Precision and Recall to illusrate that instruction-tuning of models increases the quality of the text, at a cost of a loss in diversity 3) we show how Precision and Metrics can be used to characterize models relatively to each other, without a reference dataset.

\subsection{Experimental settings}
For the models, we stress the distinction between \emph{pre-trained} models and \emph{instruction-tuned} (or \emph{chat}) models. The former are models that have been trained on a large corpus of text, and the latter have been fine-tuned with instructions and optionally aligned with human feedback. More details on the experimental settings are provided in \Cref{app:experiments}.

\paragraph{Models.} We considered the following recent LLMs:
\begin{itemize}[leftmargin=*, topsep=0pt]\setlength{\itemsep}{-0em}
    \item \textsc{Llama-2~\{7, 13, 70\}B}  and their counterparts fined-tuned and aligned with human-feedback \citep{touvron2023llama}.
    \item \vicuna \citep{vicuna}, derived from \llamaS model trained on a large open-source instruction dataset.
    \item \mistral and its instruction-tuned version \mistralchat \citep{mistral}.
    \item \pythia~\textsc{\{6.9,12\}B}~\citep{pythia}.
\end{itemize}

\paragraph{Text generation.} For all tasks, we use the model to generate 4000 samples. For WebText and Creative writings, we additionally use five different seeds for the random generator to account for the stochasticity of the generation process. We then compute the average Precision and Recall over the the different seeds.

\paragraph{WebText generation.} We run generations of a wide suits of LLMs, of sizes ranging from 7B parameters to 70B parameters, with both pre-trained and instruction-tuned models. In this setting, we make models generate texts close to their pre-training data distribution. Models are prompted with a minimal instruction of 10 words extracted from the WebText train set, and we evaluate their ability to continue the prompt. To account for the different tokenizations, input and generation lengths are constrained in terms of words, rather than in terms of tokens.

\paragraph{Biographies generation.} We evaluate the abilities of LLMs to generate Wikipedia-like biographies, given some in-context examples of biographies.  The benchmark for comparison consists of summary sections from Wikipedia pages of individuals who have been distinguished with either "Good Article" or "Featured Article" accolades. The models receive a prompt that includes an instruction followed by a varying number of in-context biography examples. Our evaluation focuses on analyzing how the quality and diversity of the generated biographies evolves as we increase the number of in-context examples provided.
\begin{figure*}[t!]
\centering
    \subfloat[Correlation between Precision, Recall, MAUVE and traditional metrics.]{
        \includegraphics[height=0.21\textwidth]{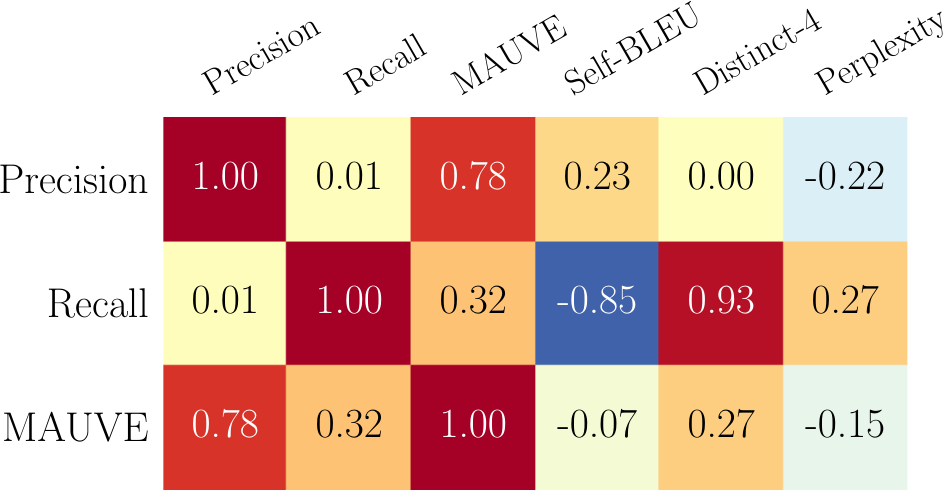}
        \label{fig:wiki-corr}}
    \hspace{0.05\textwidth}
    \subfloat[Correlation between Precision, Recall, MAUVE and the ChatGPT-3.5-turbo metrics.]{
        \includegraphics[height=0.21\textwidth]{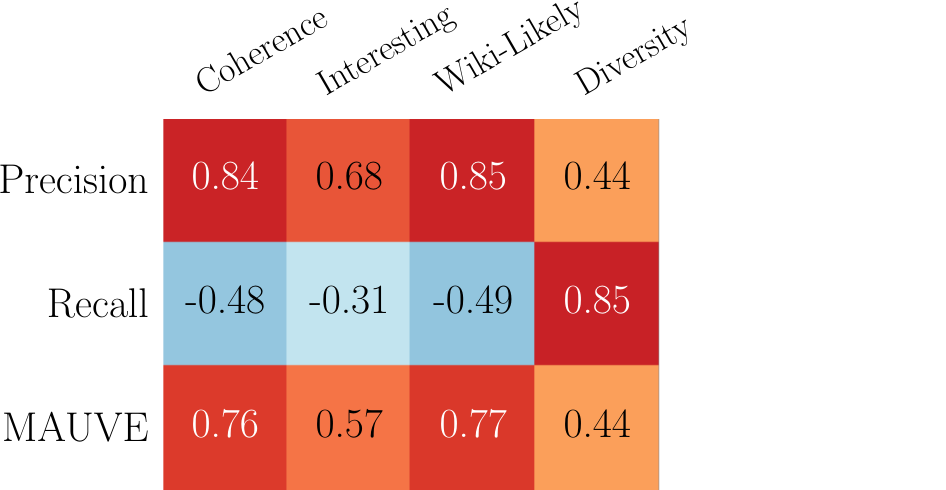}
        \label{fig:wiki-corr-gpt}}
    \caption{Correlation with Precision, Recall and MAUVE with different metrics for Wikipedia's biographies generation. Precision is on average positively correlated to quality evaluation, while Recall is positively correlated with diversity evaluation.}
\end{figure*}

\paragraph{Creative texts generation.}
We task the models with generating creative texts based on a set of 50 manually-crafted creative instructions such as \texttt{Write about a dream you had} or \texttt{Write a script for a short film}. We leverage the stochastic decoding procedure of models to generate a large number of outputs based on these 50 prompts. The specificity of this task is that there is no reference dataset available.

\subsection{Preliminary results}\label{sec:preliminary-results}

First, we present some preliminary results to illustrate the potential of our metrics.
\begin{figure*}[t!]
    \centering
    \subfloat[t][Pre-trained models.]{\includegraphics[width=0.95\textwidth]{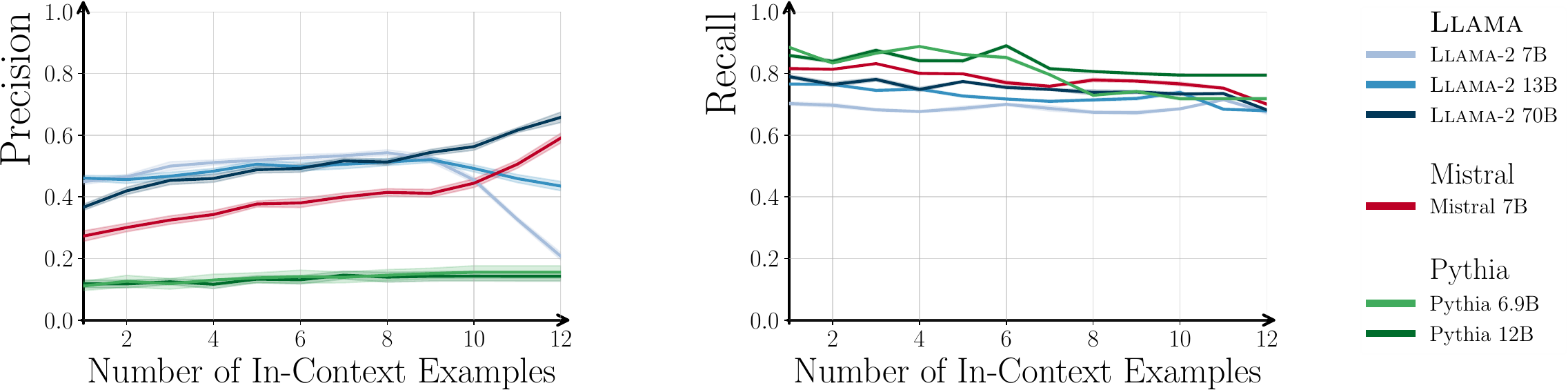}\label{fig:wiki-non-chat}}

    \subfloat[t][Instruction-tuned models.]{\includegraphics[width=0.95\textwidth]{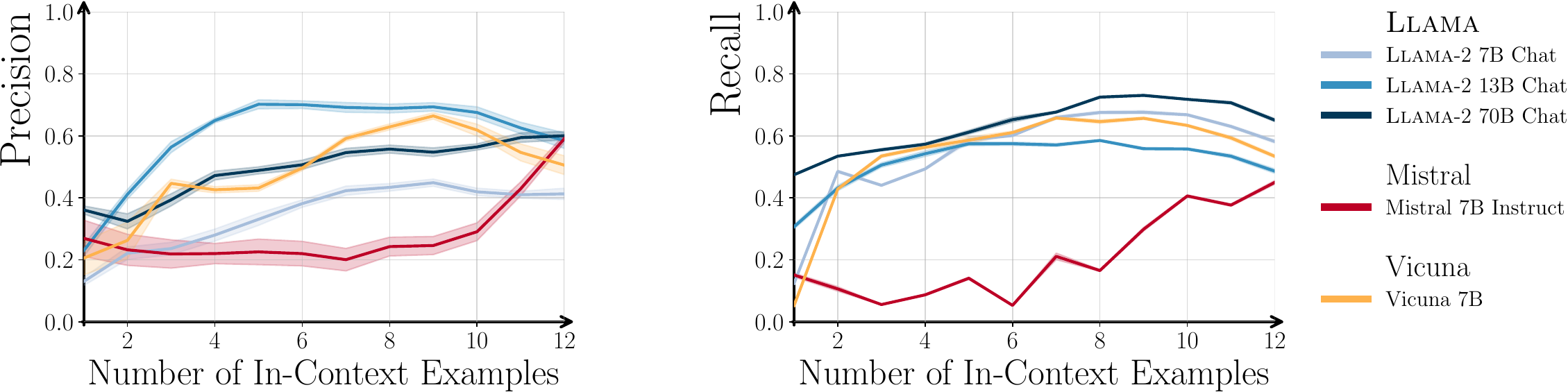}\label{fig:wiki-chat}}
    \caption{Evolution of Precision and Recall for Wikipedia's biographies generation based on the number of in-context examples. Standard deviations are depicted with transparency. Pre-trained models consistently exhibit higher Recall compared to Chat models. Precision and Recall of instruction-tuned models progressively improve until reaching a plateau after a certain number of in-context examples.}
    \label{fig:wiki-precision}
\end{figure*}

\paragraph{AG news topics modeling.} We consider a reference distribution $P$ made up of journal articles on two topics (\texttt{World, Sport}), extracted from the AG news dataset \citep{zhang_character-level_2015}. We then simulate three models. $Q_1$ corresponds to a model that generates texts only from the \texttt{World} topic. $Q_2$ generates texts from the two reference topics. $Q_3$ generates text on divergent topics compared to $P$. When considering $P$ as a reference, $Q_1$ therefore suffers from a lack of diversity and $Q_3$ from a lack of quality. The experiment is illustrated in \Cref{fig:agnews}. Precision and Recall  metrics  allow us to clearly distinguish between these cases, while, for instance, MAUVE would give the same score for both, demonstrating that our metrics can capture the trade-off between quality and diversity on a content-based level.

\paragraph{Correlation with other metrics.} \Cref{fig:wiki-corr} shows the correlation between Precision and Recall and other distribution-based metrics on the Wikipedia Biography dataset in our experiments. We observe that:
\begin{enumerate}[leftmargin=*, topsep=0pt]\setlength{\itemsep}{-0em}
    \item Recall expectedly reflects word-level diversity, as indicated by its correlation with Self-BLEU and Distinct-4.
    \item Precision shows no correlation with these metrics, but does correlate with MAUVE.
\end{enumerate}
This suggests that Precision captures the adequacy of distributions, but not diversity, essentially indicating the quality of the generated texts. A complementary analysis of MAUVE on the WebText dataset is available in \Cref{app:webtext}.

\subsection{Evaluation of open-ended generation}\label{sec:use-cases:open-ended}

\paragraph{Instruction-tuned models are more precise and less diverse than pre-trained models.} \Cref{fig:webtext-PR} shows the Precision / Recall graph of various models. Instruction tuned models have a higher Precision and a lower Recall than their pre-trained counterparts. This is consistent with the expected effect of human preference alignment and instruction tuning: models are encouraged to generate more "human-like" texts, which do not encompass any notion of diversity.
\paragraph{Larger models are more diverse.} The Recall is consistently better for larger models. Larger models have a better expressive power, and this diversity is reflected by their higher Recall.

\subsection{Generating Biographies.}

The results are presented in \Cref{fig:wiki-non-chat,fig:wiki-chat}. We plot the Precision and the Recall as a function of the number of in-context examples.
\paragraph{The number of in-context examples makes the model more precise.} For both pre-trained and instruction-tuned models, the Precision increases with the number of in-context examples. This is consistent with the intuition that several in-context examples help the model generate texts that are closer to the expected distribution.
\paragraph{Increasing the number of in-context examples make "chat" models more diverse.} \Cref{fig:wiki-chat} confirms our previous findings that instruction-tuned models are less diverse than their pre-trained counterparts. However, by increasing the number of in-context examples, we observe a strong regain in diversity: chat models leverage the diversity in the prompt to generate more diverse texts.
\paragraph{Chat models plateau.} After a few numbers of in-context examples, we observe that both the Precision and the Recall plateau. This suggests that the model has captured the distribution of the reference datasets after a few examples and that adding more examples does not bring additional information.

\begin{table*}[ht!]
    \centering
    \subfloat[Estimation of overlaps of the distribution of \textsc{Llama}-2 70B Chat.]{\label{tab:crea-llama70b}
        \small \centering
        \begin{tabular}{@{}lccccc@{}}
            \toprule
                      & \textbf{\textsc{Llama}-2 70B Chat} & \textsc{Llama}-2 13B Chat & \textsc{Llama}-2 7B Chat & Mistral 7B Instruct & Vicuna 7B      \\ \midrule
            Precision & $\mathbf{1.00 \pm 0.00}$           & $0.95\pm 0.01$            & $0.94\pm 0.01$           & $0.99\pm 0.01$      & $0.93\pm 0.01$ \\

            Recall    & $\mathbf{1.00 \pm 0.00}$           & $0.96\pm 0.01$            & $0.92\pm 0.01$           & $0.83\pm 0.08$      & $0.96\pm 0.02$ \\
            \bottomrule
        \end{tabular}}\\
    \subfloat[Estimation of overlaps of the distribution of \mistralchat.]{\label{tab:crea-mistral}
        \small \centering
        \begin{tabular}{@{}lccccc@{}}
            \toprule
                      & \textbf{Mistral 7B Instruct} & \textsc{Llama}-2 70B Chat & \textsc{Llama}-2 13B Chat & \textsc{Llama}-2 7B Chat & Vicuna 7B      \\ \midrule
            Precision & $\mathbf{1.00 \pm 0.00}$     & $0.82\pm 0.07$            & $0.77\pm 0.08$            & $0.78\pm 0.09$           & $0.67\pm 0.01$ \\
            Recall    & $\mathbf{1.00 \pm 0.00}$     & $0.98\pm 0.01$            & $0.93\pm 0.01$            & $0.91\pm 0.01$           & $0.95\pm 0.01$ \\
            \bottomrule
        \end{tabular}
    }
    \caption{Relative comparison of models based on Precision and Recall on the Creative text generation task. The texts generated by the first models are used as a reference against which are compared the others. The values are averaged over 5 generation seeds and the standard deviation is given. The model used as a reference distribution is in \textbf{bold}.}
\end{table*}

\subsection{Creative text generation}

For this task, no reference dataset is available. Still, our novel Precision and Recall metrics offer a means to compare models, where conventional methods fall short.

\paragraph{Estimation of overlaps of the distribution of models.} Since our metrics Precision and Recall corresponds to different support overlap, they can be used to analyze models comparatively.

\paragraph{Models of the same family share close distributions but size induces some differences.} \Cref{tab:crea-llama70b} shows that compared to the reference \llamaXchat, Precision and Recall of \llamaSchat and \llamaMchat increases with their size.

\paragraph{Intepreting distributions overlap.} Compared to \llamaXchat, \mistralchat has an almost perfect Precision but suffers from a lower Recall. Geometrically, this means that its distribution fits almost perfectly in the support of \llamaXchat, but does not cover it totally.
\Cref{tab:crea-mistral} illustrates even further this phenomena: compared to \mistralchat, every model has a quite low precision but a high recall. Geometrically, the \textsc{Llama-2}-based models cover a wide portion of the support of \mistralchat, but also assigns mass to points outside this distribution.

\subsection{Correlation with human-like preferences}

Finally, we evaluated how our proposed metrics relate to human evaluation. Since one of our evaluation involve significantly long inputs, we conducted a mock-human evaluation using GPT-3.5-turbo. To evaluate the quality, we ask GPT-3.5-turbo to rate our generated biographies on three dimensions: coherence, insightfulness and Wikipedia-likeness. We also experimented with a proxy of diversity, by asking GPT-3.5-turbo to rate the similarity of batches of 10 texts (i.e., low score if there are many similar texts). More details regarding the experimental setup are given in \Cref{app:gpt-eval}.

\paragraph{Precision Correlates with GPT-3.5 quality evaluation.} Models with high precision generate texts that align closely with the reference distribution, which consists of high-quality Wikipedia articles. Consequently, higher precision typically corresponds to better GPT-3.5 scores, as illustrated in \Cref{fig:wiki-precision}. Note that instruction-tuned models, which are tuned similarly to GPT-3.5, were excluded from the correlation analysis with GPT-3.5 quality scores due to their uniformly high scores, relatively to pre-trained models.

\paragraph{Recall correlates with human-like diversity evaluation.} As depicted on \Cref{fig:wiki-precision}, we found a positive correlation between our introduced Recall and human-like evaluation of diversity. This confirms our findings regarding the generated contents of LLMs.

\section{Conclusion}
We introduced Precision and Recall as automated metrics for comparing generative model outputs against a reference distribution. These metrics serve as independent indicators of the quality and diversity of the generated text. In challenging scenarios where evaluation protocols are scarce, such as tasks with high generation complexity, we demonstrate the efficacy of our metrics in assessing the quality and diversity of model output. Furthermore, our metrics revealed nuanced model behaviors that lie beyond the scope of traditional evaluation benchmarks. We believe that our work contributes to advancing the evaluation of generative models, offering valuable insights to better understand their capabilities, and aiding in the development of more refined evaluation protocols within the community.
\section{Limitations}
Our study proposes to adapt the framework of \citet{kynkaanniemi_improved_2019} from image to text generation. However, limitations of the original image-based framework, as highlighted by \citet{naeem_reliable_2020} and \citet{kim_toppr_2023}, may also apply to our adaptation to NLP. In particular, the sensitivity to outliers and the generalizability of these metrics across different domains warrant further investigation.

The metrics proposed in our study are based on an auxiliary model, specifically GPT-2\textsubscript{LARGE} embeddings, to rate the quality of text generations. Although the metrics tend to be correlated across models as shown in \Cref{app:other-embeddings}, the choice of embeddings may still significantly impact the performance and applicability of these metrics. Future research could explore the effects of alternative embedding models on the Precision and Recall measurements in text generation tasks.

We present these metrics as valuable tools for evaluating a model's open-ended generation capabilities. However, to comprehensively assess LLMs across specific dimensions and tasks, these metrics should be supplemented by other approaches. In particular, the relationship with traditional human evaluation needs to be investigated more thoroughly in future research. Although we performed a mock human evaluation using GPT-3.5-turbo, this is not a complete assessment of how these metrics compare to the human evaluation.

We restricted our analysis to English language on most popular LLMs.
\section{Ethical considerations}
Precision and Recall are metrics that rely heavily on the distribution of the data they are tested against. If the data contain biases, these metrics may favor models that perpetuate those biases. This issue is crucial because it can lead to unfair outcomes, particularly for underrepresented groups, emphasizing the need for a well-designed reference dataset.

To address this problem, we propose a practical solution: calculate Precision and Recall for specific groups within the data, such as different genders or ethnic minorities. This approach helps to ensure that the model is equally accurate and fair between all groups. Our objective is to prevent models from overlooking or misrepresenting any group. Further investigations should focus on this consideration.

\section{Acknowledgements}
This work has been partly funded through project ACDC ANR-21-CE23-0007.
This project was provided with computing AI and storage resources by GENCI at IDRIS thanks to the grants 20XX-AD011014022R1, 20XX-AD011014053 and 20XX-A0151014627 on the supercomputer Jean Zay's V100/A100 partition.
\bibliography{custom,references}

\appendix
\renewcommand\thefigure{\thesection.\arabic{figure}}
\counterwithin{figure}{section}
\counterwithin{table}{section}
\onecolumn
\section{Definition of Precision-Recall Curves}
\label{app:prcurves}
In this section, we introduce a mathematical tool called the PR-Curve, which is used to evaluate generative models. PR-Curves are closely related to the MAUVE score. We will then list the main differences between PR-Curves and the MAUVE score to highlight the limitations of the latter. Following this, we will demonstrate with a straightforward example that the MAUVE score struggles to distinguish between a lack of diversity and a lack of quality. This difficulty can be explained by some of the distinguishing points between PR-Curves and the MAUVE score.
\subsection{Definition}\label{sec:prcurvedef}

Recognizing the need to distinguish between quality and diversity, and drawing a parallel with the concepts of precision and recall from classification, \citet{sajjadi_assessing_2018} introduced a new definition of Precision and Recall for discrete distributions.

\begin{figure}[H]
  \begin{minipage}{0.47\linewidth}
    \begin{definition}[Precision-Recall trade-off \cite{sajjadi_assessing_2018}]
      A distribution $Q$ has a precision $\alpha$ and a recall $\beta$ with respect to a distribution $P$ if there exists distributions $\mu$, $\nu_P$ and $\nu_Q$ such that:
      \begin{subequations}
        \begin{align}
          P = \alpha \mu & +(1- \alpha )\nu_P \\ Q = \beta \mu& + (1-\beta) \nu_Q.
        \end{align}
        \label{eq:pq_sajjadi}
      \end{subequations}

    \end{definition}
  \end{minipage}\hfill
  \begin{minipage}{0.47\linewidth}
    \begin{figure}[H]
      \centering
      \includegraphics[width=\linewidth]{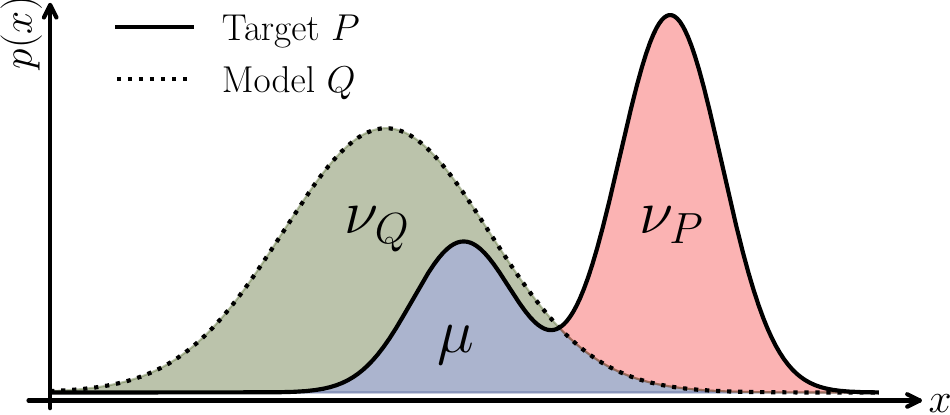}
      \caption{Precision-Recall trade-off illustration}
      \label{fig:pr_tradeoff}
    \end{figure}
  \end{minipage}
\end{figure}

This definition is illustrated in \Cref{fig:pr_tradeoff}.
The distribution $\nu_P$ represents the part of $P$ that cannot be generated by $\mu$ and thus $Q$, and $\nu_Q$ represents the part of $Q$ that should not be generated by $\mu$. The set of all attainable couples $(\alpha, \beta)$ such that there is a distribution $\mu$ from \Cref{eq:pq_sajjadi} defines various \textit{ trade-offs}. The Precision Recall Curve can then be defined as the set of all Pareto-optimal trade-offs. More recently, \citet{simon_revisiting_2019} has extended the definition for any continuous distribution $P$ and $Q$:
\begin{theorem}[PR-Curve \cite{simon_revisiting_2019}]
  The set of Pareto-optimal trade-offs, that is, the PR-Curve, can be represented as the set of points $(\alpha_\lambda, \beta_\lambda)_{\{\lambda\in [0, \infty]\}}\in[0,1]^2$ such that:
  \begin{align*}
    \begin{cases}
      \alpha_\lambda = \E_Q\left[\min\left(1, \lambda\frac{p(\vx)}{q(\vx)}\right)\right], \\[5pt]
      \beta_\lambda = \E_P\left[\min\left(1, \frac{1}{\lambda}\frac{q(\vx)}{p(\vx)}\right)\right].
    \end{cases}
  \end{align*}
\end{theorem}

\begin{figure}
  \centering
  \subfloat[Reference distribution $P$.]{\includegraphics[width=0.18\textwidth]{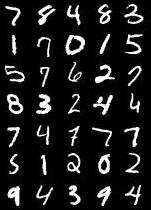}\label{app:fig:mnist_dist}}\hspace{0.02\textwidth}
  \subfloat[Distribution $Q_1$.]{\includegraphics[width=0.18\textwidth]{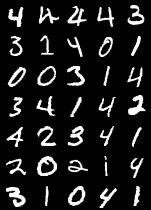}\label{app:fig:q1_dist}}\hspace{0.02\textwidth}
  \subfloat[Distribution $Q_2$.]{\includegraphics[width=0.18\textwidth]{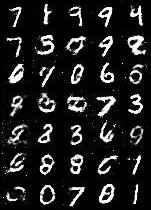}\label{app:fig:q2_dist}}\hspace{0.02\textwidth}
  \subfloat[PR-Curves of $Q_1$ and $Q_2$ w.r.t $P$.]{\includegraphics[width=0.35\textwidth]{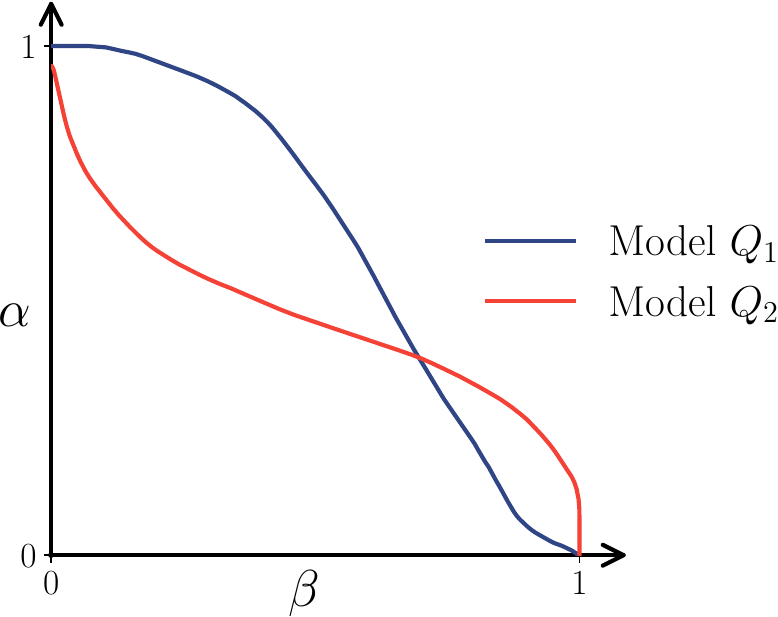}\label{app:fig:prs_mnist}}
  \caption{PR-Curves for the MNIST dataset. The target distribution $P$ is shown in (a), and the distributions $Q_1$ and $Q_2$ are shown in (b) and (c) respectively. The PR-Curves of $Q_1$ and $Q_2$ are shown in (d). The PR-Curve of $Q_1$ indicates high quality and low diversity and the PR-Curve of $Q_2$  indicates low quality but high diversity.}
  \label{app:fig:pr_mnist}
\end{figure}

Intuitively, for any given trade-off parameter $\lambda\in[0, +\infty]$, the precision $\alpha_\lambda$ diminishes when $q(\vx)$ is smaller than $\lambda p(\vx)$ at certain points $\vx$, indicating overestimation of regions by the distribution $Q$. On the contrary, the recall $\beta_\lambda$ declines when $p(\vx)$ is smaller than $\lambda q(\vx)$ for specific points $\vx$, suggesting underestimation of regions by $Q$. More concretely, an example of PR-Curves is depicted in \Cref{app:fig:pr_mnist}. \Cref{app:fig:mnist_dist} shows samples drawn from the reference distribution $P$. Here, $Q_1$ represents a distribution capable of generating high-quality images, albeit only for the first 5 digits, while $Q_2$ represents a distribution that generates noisy images on all labels. The PR-Curve of $Q_1$ surpasses that of $Q_2$ for high values of $\lambda$ (top left), indicating the superior quality of $Q_1$. In contrast, for low values of $\lambda$ (lower right), the PR-Curve of $Q_2$ exceeds that of $Q_1$, indicating greater diversity in $Q_2$.

For a two-number summary of these curves, \citet{sajjadi_assessing_2018} has proposed the $F_\gamma$-scores: $\mathrm{F}_\gamma=\max_\lambda(1+\gamma)^2\alpha_\lambda\beta_\lambda/(\gamma^2\alpha_\lambda+\beta_\lambda)$. The $\mathrm{F}_{1/8}$ is employed to assess the quality of the model, and $\mathrm{F}_{8}$ score is used to assess diversity. Abusing the notation, the scores $\mathrm{F}_{1/8}$ and $\mathrm{F}_{8}$ are often referred as Precision and Recall. In practice, the PR-Curves are estimated by sampling from the distributions $P$ and $Q$, and by performing density estimation on the embeddings on the samples. There are several ways to estimate the PR-Curves, including a $k$-means estimator \citep{sajjadi_assessing_2018} or a discriminator method \citep{simon_revisiting_2019}.

\paragraph{Connection with PR-Curves and Precision/Recall: } In this paper, we have used another two-number summary of the PR-Curve introduced by \citet{kynkaanniemi_improved_2019}: the maximum precision $\alpha_\infty$ and the maximum recall $\beta_0$. As a matter of fact, one can show that:
\begin{align*}
  \alpha_\infty = \max_\lambda \alpha_\lambda = Q(\Supp(P)) \quad \text{and} \quad \beta_0 = \max_\lambda \beta_\lambda = P(\Supp(Q)).
\end{align*}

These two values encapsulate the definition of support provided by \citet{kynkaanniemi_improved_2019}. The first value, $\beta_0 = P(\Supp(Q))$, measures the proportion of $P$ covered by the support of $Q$. The second value, $\alpha_\infty = Q(\Supp(P))$, measures the proportion of $Q$ covered by the support of $P$. The primary practical advantage is that these two values can be computed without performing a density estimation. As detailed in \Cref{sec:prdef}, we use a $k$-nearest neighbors estimator to estimate these values.

%For more refine comparison, the PR-Curve can be hard to interpret. The extreme values of the PR-Curve are encapsulating the definition of support given by \citet{kynkaanniemi_improved_2019}: $\beta_0 = P(\Supp(Q))$ assess the mass of $P$ generated by $Q$ with any density and $\alpha_\infty = Q(\Supp(P))$ assess the mass of $Q$ generates $P$. And, as we have said, if the support of the distribution matches, then $\alpha_\infty = \beta_0 = 1$, and these values become useless for model evaluation.
\begin{figure}[b!]
  \centering
  \includegraphics[width=0.95\textwidth]{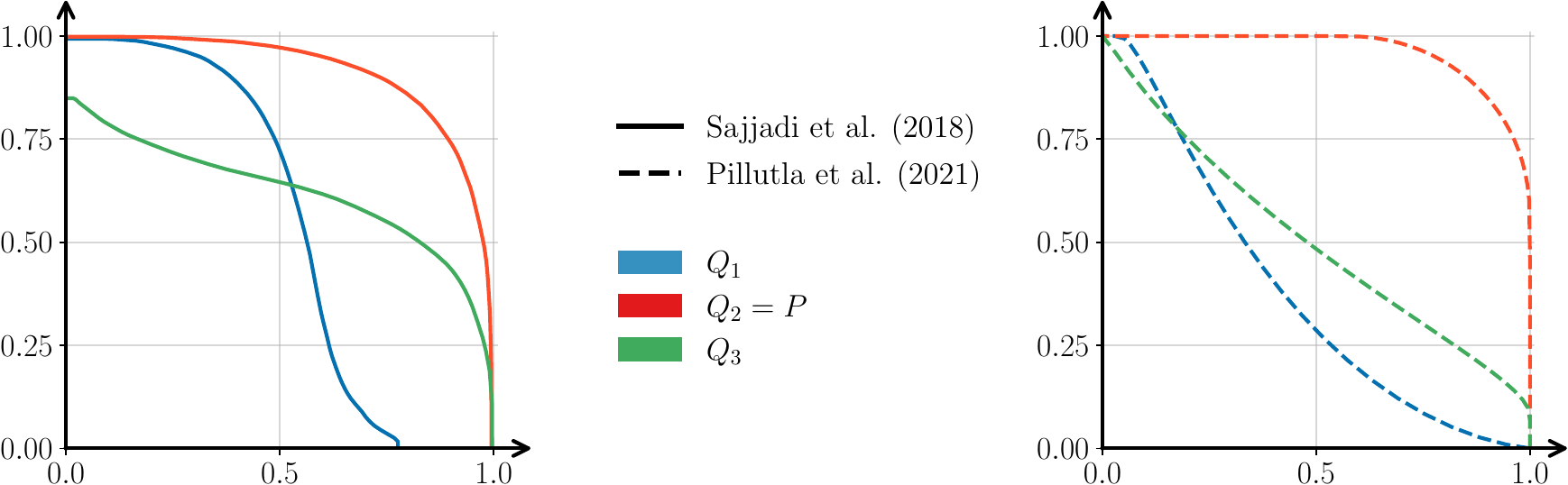}
  \caption{Comparison of the PR-Curve from \citep{sajjadi_assessing_2018} (left) and the divergence frontier used to compute MAUVE score \citep{pillutla_mauve_2021} for the AG News dataset. Visually, MAUVE's curve struggle to distinguish between a lack of diversity and a lack of quality, while the PR-Curves on the left provide a finer evaluation.}
  \label{app:fig:agnewsprcurves}
\end{figure}
\subsection{MAUVE score}\label{app:mauve}
\paragraph{MAUVE score definition.} The MAUVE score \citep{pillutla_mauve_2021} is a measure of discrepancy between two distributions $P$ and $Q$ in the embedding space. The MAUVE score is defined as the area under the divergence frontier between $P$ and $Q$. The divergence frontier is defined as the set of points $(\alpha_\pi, \beta_\pi)$ for $\pi\in[0, 1]$:
\begin{align*}
  \begin{cases}
    \alpha_\pi = \exp\left(-\KL(Q\Vert \pi P + (1-\pi)Q)\right) \\
    \beta_\pi = \exp\left(-\KL(P\Vert \pi P + (1-\pi)Q)\right).
  \end{cases}
\end{align*}
In other words, the MAUVE score is a one-number summary of another quality-diversity curve.

\paragraph{Connection between MAUVE and PR-Curves.} The MAUVE score is based on a curve different from PR-Curves, and there is no straightforward connection between the two. But, from a practical point of view, both curves are estimated using density estimation on the embeddings of samples from $P$ and $Q$. \citet{pillutla_mauve_2021} used a $k$-means estimator to estimate the divergence frontier and then computed the area under the curve to obtain the MAUVE score.

Since both curves are based on density estimation, we can compare them using the same dataset, embeddings, and density estimation method. This allows us to highlight the limitations of the divergence frontier used in the MAUVE score. In our AG News experiment, fully described in \Cref{sec:preliminary-results}, we exemplify these limitations. Specifically, \Cref{app:fig:agnewsprcurves} shows that the underlying curves used for computing the MAUVE score struggle to differentiate the discrepancies highlighted in \Cref{fig:agnews}, while the original PR-Curves from \citet{sajjadi_assessing_2018} can.

\begin{figure}[H]
  \centering
  \includegraphics[width=0.95\textwidth]{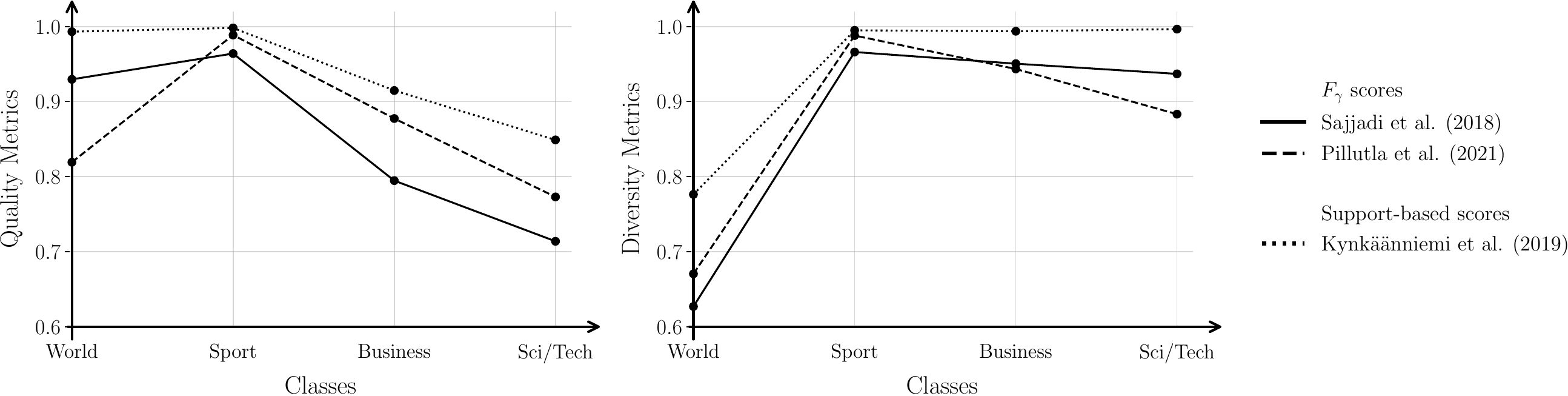}
  \caption{The $\mathrm{F}_{1/8}$ and $\mathrm{F}_{8}$ scores for the AG News dataset. The reference consists of texts of \texttt{World, Sport} topics. We plot the evolution of the scores with respect to this reference as we add topics to the candidate distribution. Scores from \citet{pillutla_mauve_2021} do not identify the lack of recall of when topics are missing from the candidate distribution.}
  \label{app:fig:agnewsFgamma}
\end{figure}

When adapting the $\mathrm{F}_{1/8}$ and $\mathrm{F}_{8}$ scores to the MAUVE curves and plotting the results in \Cref{app:fig:agnewsFgamma}, we find that these adapted scores are substantially less meaningful than those from \citet{sajjadi_assessing_2018}. Furthermore, the definitions of Precision and Recall we used, based on support-based metrics \citep{kynkaanniemi_improved_2019}, appear to be more informative than the $\mathrm{F}_{1/8}$ and $\mathrm{F}_{8}$ scores.

\begin{figure}[H]
  \centering
  \includegraphics[width=0.7\textwidth]{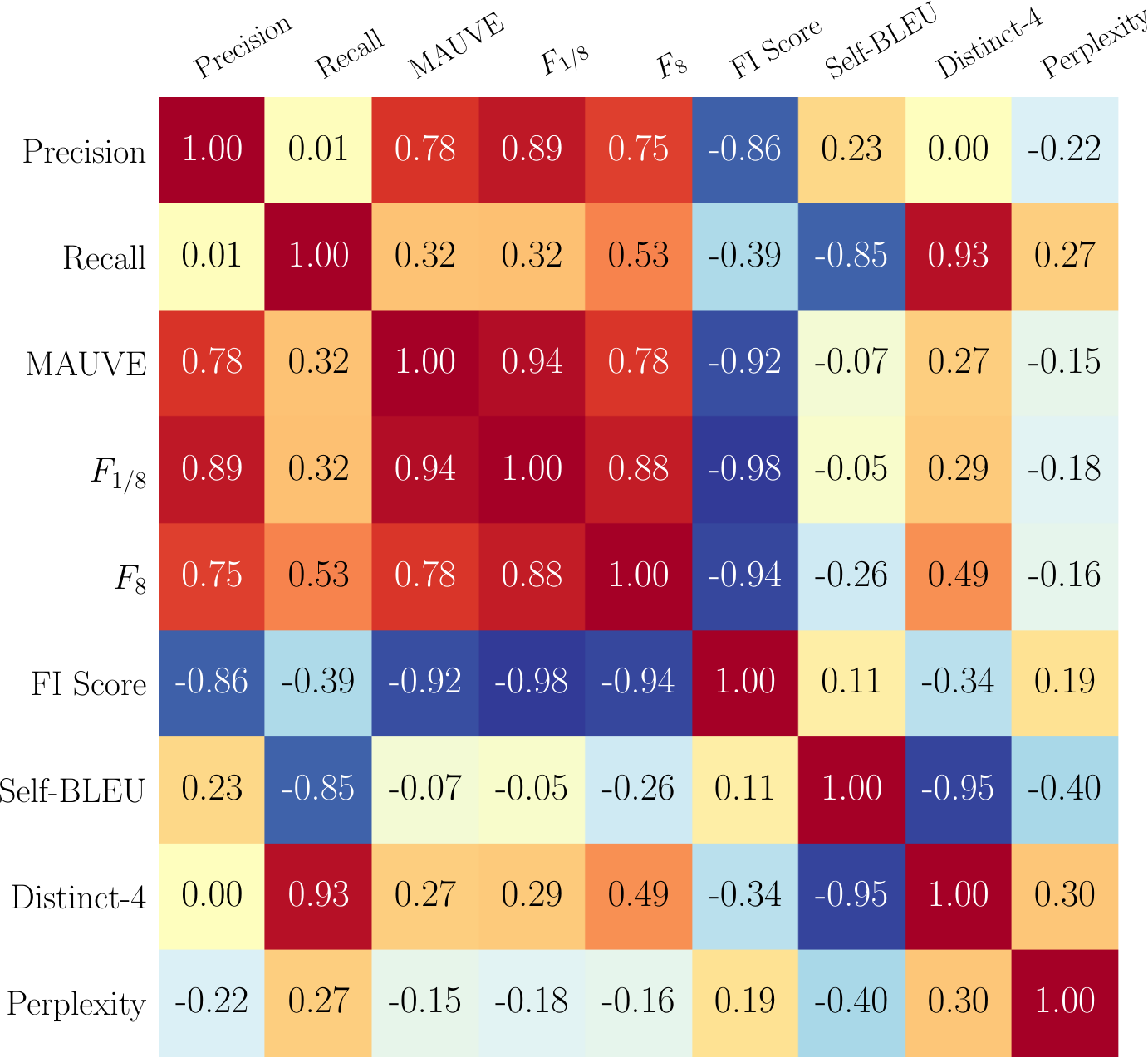}
  \caption{Correlation between Precision, Recall, MAUVE \cite{pillutla_mauve_2021} and the $\mathrm{F}_{1/8}$ and $\mathrm{F}_{8}$ scores \citet{sajjadi_assessing_2018} on the Wikipedia's biographies experiment. $\mathrm{F}_{1/8}$ and $\mathrm{F}_{8}$ correlate more with Precision than with Recall and strongly correlate with each other.}
  \label{app:fig:correlation_sajjadi}
\end{figure}

\paragraph{Difference between Precision/Recall and MAUVE.} As detailed in the previous paragraph, the MAUVE score is a single-number summary of a divergence frontier, whereas the PR-Curves provide a two-number summary of the precision-recall curve. Consequently, in addition to the fact that the divergence frontier is less capable of distinguishing between quality and diversity, MAUVE, being a single metric, is inherently less informative than Precision and Recall. In \Cref{app:fig:correlation_sajjadi}, we examine the correlations between Precision, Recall, the MAUVE score, and the $\mathrm{F}_{1/8}$ and $\mathrm{F}_{8}$ scores on Wikipedia biographies, as detailed in \Cref{sec:use-cases:open-ended}. We observe that the $\mathrm{F}_{1/8}$ and $\mathrm{F}_{8}$ scores correlate more with Precision than with Recall and strongly correlate with each other. This strongly indicates that the MAUVE score cannot effectively differentiate between the quality and diversity of models.

To conclude, MAUVE and our metrics are all distribution-based metrics that rely on discretized approximations of both the target distribution \(P\) and the model's distribution \(Q\), computed using embedding functions. However, we can identify three reasons why MAUVE differs from the metrics introduced in this paper:

\begin{itemize}
  \item MAUVE is a one-number summary of a divergence frontier, whereas Precision and Recall provide a two-number summary of the PR-curve.
  \item MAUVE is based on the divergence frontier, which has been experimentally shown to be less informative than the PR-Curves.
  \item MAUVE uses the divergence frontier estimated with $k$-means, while Precision and Recall are estimated with a $k$-nearest neighbors estimator.
\end{itemize}

\section{Experiments}\label{app:experiments}

\subsection{Other embedding functions}\label{app:other-embeddings}

We evaluated the biographies generated using different embedding models. We computed the correlation between the embeddings $\text{GPT-2}_\text{LARGE}$ and the other embeddings and reported the results in \Cref{tab:correlation_embeddings}. All Precision and Recall scores are positively correlated, demonstrating the robustness of our method. Notably, stronger correlations are observed between the decoder-only models, likely due to their pre-training objectives, which make them well-suited for evaluating open-ended text generation.

\begin{table}[h!]
    \centering
    \resizebox{\textwidth}{!}{
        \begin{tabular}{lcccccc}
            \hline
                      & GPT-2-large & GPT-2-xl & OPT-1.3b            & Gemma-2b                  & BART-large                  & RoBERTa-large         \\
                      &             &          & \cite{zhang2022opt} & \cite{gemmateam2024gemma} & \cite{lewis-etal-2020-bart} & \cite{liu2019roberta} \\
            \hline
            Precision & 1.0         & 0.99     & 0.98                & 0.77                      & 0.70                        & 0.85                  \\
            Recall    & 1.0         & 0.99     & 0.89                & 0.86                      & 0.77                        & 0.67                  \\
            \hline
        \end{tabular}
    }
    \caption{Correlation between different embedding models.}
    \label{tab:correlation_embeddings}
\end{table}

\subsection{GPT-3.5-turbo evaluation}\label{app:gpt-eval}
Here we describe the complete evaluation setup we used for GP-T3.5-turbo evaluation.

\paragraph{Evaluation dimensions and instructions.} Below is a description of the evaluation dimensions we considered for both the three quality dimensions and diversity.
\begin{itemize}
    \item \textbf{Coherence:} GPT-3.5-turbo was asked to rate the ease of reading and understanding the text on a scale from 1 (difficult to read) to 5 (very easy to read).
    \item \textbf{Insightfulness:} The insightful or interesting nature of the biography was rated on a scale from 1 (not interesting) to 5 (highly interesting).
    \item \textbf{Plausibility:} This dimension assessed how likely the text is to be a biography from Wikipedia, with ratings from 1 (not likely) to 5 (very likely).
    \item \textbf{Diversity:} This dimension evaluated the variation among biographies, assessing if they differ in terms of names, professions, etc., with a score of 0 indicating similarity and 1 indicating diversity. This is evaluated on batches of 10 texts.
\end{itemize}

\paragraph{Evaluation process.} Below are the instruction and prompt given to GPT-3.5-turbo.

\begin{itemize}
    \item \textbf{For assessing quality:}
          \begin{itemize}
              \item \textbf{Instruction:} "You will be asked to rate a Wikipedia biography based on coherence, insightfulness, and likelihood on a scale from 1 to 5. Please provide three numerical answers using the template in JSON (\texttt{"score\_coherence": score, "score\_insightful": score, "score\_likely": score})."
              \item \textbf{Prompt:} "On a scale of 1 to 5, rate the following biography on three criteria:
                    \begin{itemize}
                        \item \textbf{Coherence:} How easy is it to read and understand the text? (1 for difficult to read, 5 for very easy to read)
                        \item \textbf{Insightfulness:} How insightful or interesting is the biography? (1 for not interesting, 5 for highly interesting)
                        \item \textbf{Probability:} How likely is this text a biography from Wikipedia? (1 for not likely, 5 for very likely)
                    \end{itemize}
                    Biography Text: \{text\}

                    Scores:"
          \end{itemize}
    \item \textbf{For assessing diversity:}
          \begin{itemize}
              \item \textbf{Instruction:} "You will be asked to rate the similarity between several biographies. Give a score of 0 if some biographies are similar in terms of names, profession, etc., and give a score of 1 if they are all different. Provide the answer in JSON (\texttt{"similarity\_score": score})."
              \item \textbf{Prompt:} "Rate the similarity of the following texts:
                    \begin{itemize}
                        \item[] \{concatenated\_texts\}
                    \end{itemize}
                    Similarity Score:"
          \end{itemize}
\end{itemize}

\subsection{Implementation details}\label{app:implementation}
This section provides full details on the experimental protocol adopted in this paper.
\paragraph{Texts input length.} All tested models do not share the same tokenization. To avoid a potential bias due to different generations' lengths, input and output length constraints are expressed in terms of words rather than tokens. The average number of tokens per word has been calculated for each model on the train set of the WebText dataset\footnote{Available at \url{https://github.com/openai/gpt-2-output-dataset}} and reported in \Cref{tab:tokens-per-word}.

\paragraph{Hardware and software setup.} Generations were conducted on either A100 GPUs with 80Gb memory or on V100 with 32Gb memory. Models were run with a bfloat16 precision on A100 and with float32 precisions on V100, except for the \textsc{Pythia} models, which have been ran with float16 mixed precision, following the original setup \citep{pythia}. For models with $\approx 7B$ parameters,  generating 4000 samples takes approximately 5 hours on 2 A100s, it takes about 10 hours for models with $\approx 13B$ parameters on 2 A100, and about 40 hours for models with $\approx 70B$ parameters on 4 A100. Generating the samples required approximately 4200 GPUs hours.
Code is built on PyTorch \citep{paszke_pytorch_2019}, with the HuggingFace library \citep{wolf_huggingfaces_2020}.

\paragraph{Computational cost of Precision and Recall.} Since Precision and Recall are based on a $k$-NN computations, computing these metrics is especially fast, even across large datasets. Once the features are computed, computing Precision and Recall takes less than 5 seconds on a standard laptop for 4000 samples, instead computing MAUVE takes approximately 40s.

\paragraph{Other metrics implementations.} For computing SelfBLEU, Disctinct-N and MAUVE we relied on the official implementations found in \url{https://github.com/koadman/mauve}, or \url{https://github.com/krishnap25/mauve-experiments}.

\paragraph{Implementations.} The code used to run and evaluate the generations, as well as the exact datasets are available at \url{https://github.com/AlexVerine/pr-4-llm}.

\begin{table}[h]
  \centering
  \begin{tabular}{lc}
    \toprule
    \textbf{Model}   & \textbf{Average Tokens per Word} \\
    \midrule
    \textsc{Mistral} & 1.313                            \\
    \textsc{Pythia}  & 1.183                            \\
    \textsc{Llama-2} & 1.360                            \\
    \textsc{Vicuna}  & 1.360                            \\
    \bottomrule
  \end{tabular}
  \caption{Average number of tokens per word for different families of models. \textsc{Vicuna} is based on \textsc{Llama-2}.}
  \label{tab:tokens-per-word}
\end{table}

\subsection{WebText Generation}
\label{app:webtext}

\paragraph{Data.} Reference dataset is extracted from the official OpenAI WebText dataset at \url{https://github.com/openai/gpt-2-output-dataset}, under a MIT license and is composed of articles extracted from WEB urls. We extracted 15k samples from this datasets for our experiments.

\paragraph{Input formatting.} Non-chat models are simply prompted with the first 10 words of random WebText articles. For chat models, they are explicitly asked to continue the prompt. We input the last 10th word out of the instruction tokens to avoid idiomatic expressions such as \textit{Sure! Here is \ldots} and ensure fair comparison between models. The different prompts are illustrated in \Cref{tab:webtext-input-template}.

\begin{table}[htbp]
    \centering
    \caption{Input Formats for Chat and Non-Chat Models}
    \label{tab:webtext-input-template}
    \begin{tabular}{p{4cm}p{9cm}}
        \toprule
        \textbf{Model}  & \textbf{WebText input Template}                \\
        \midrule
        \textsc{Llama-2 Chat, Mistral Chat}
                        & \verb|[INST] Continue the following text:|     \\
                        & \verb|{{first 9 words}} [/INST] {{10th word}}| \\
        \midrule
        \textsc{Vicuna} & \verb|USER: Continue the following text:|      \\
                        & \verb|{{first 9 words}}|                       \\
                        & \verb|ASSISTANT: {{10th word}}|                \\
        \midrule
        Non-chat models & \verb|{{first 10 words}}|                      \\
        \bottomrule
    \end{tabular}
\end{table}

\paragraph{Generation setup.} For this task, we used the default recommended generation setup, i.e. nucleus sampling for all models, with parameters displayed in \Cref{tab:generation_parameters}. \textsc{Llama-2} default parameters are from the official repository \url{https://github.com/facebookresearch/llama/blob/main/example_text_completion.py} and \textsc{Mistral}'s ones from their official API \url{https://docs.mistral.ai/api/}. For \textsc{Pythia}'s models, we did not find any mention of recommenced generation parameters and hence used a temperature and nucleus p of $1.0$.

\begin{table}[htbp]
    \centering
    \caption{Generation Parameters Summary}
    \label{tab:generation_parameters}
    \begin{tabular}{@{}lllll@{}}
        \toprule
        \textbf{Model} & \textbf{Max New Tokens} & \textbf{Nucleus P} & \textbf{Temperature} & \textbf{Repetition Penalty} \\
        \midrule
        Llama2, Vicuna & 448                     & 0.9                & 0.6                  & 1.18                        \\
        Mistral        & 432                     & 1.0                & 0.7                  & 1.18                        \\
        Pythia         & 390                     & 1.0                & 1.0                  & 1.18                        \\
        \bottomrule
    \end{tabular}
\end{table}

\begin{figure}
    \centering
    \subfloat[Evolution of the metrics Precision and Recall as the number of samples $N$ increases. The shaded area represents the standard deviation computed over 5 random seeds for the set ouputs. The averages reach a plateau around $N=3000$.]{\includegraphics[width=0.48\textwidth]{images/diversity_numsamples.pdf}\label{app:fig:diversity_numsamples}}\hfill
    \subfloat[Evolution of the metrics Precision and Recall as the parameter $k$ increases. The shaded area represents the standard deviation computed over 5 random seeds for the set ouputs.]{\includegraphics[width=0.48\textwidth]{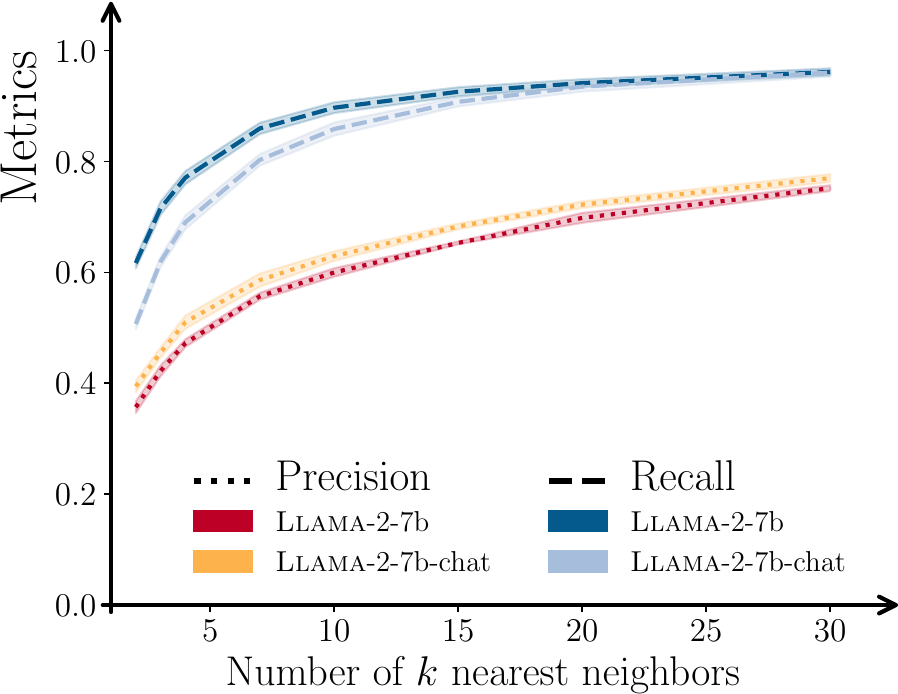}\label{app:fig:diversity_k}}
    \caption{Evolution of the metrics Precision and Recall as the number of samples $N$ and the parameter $k$ increases.}
\end{figure}

\paragraph{Evaluation setup.}
To determine the optimal number of samples and the parameter $k$ for estimating Precision and Recall, we analyze the behavior of these metrics with varying $N$ and $k$. We employ the WebText generation task with \llamaSchat as a reference. Specifically, we generate $10,000$ samples per seed and compute the average Precision and Recall for $N$ ranging from $100$ to $10,000$, and $k$ ranging from $1$ to $30$. The trends of Precision and Recall with respect to the number of samples $N$, while $k$ is fixed at $4$, are depicted in \Cref{app:fig:diversity_numsamples}. Conversely, the trends of Precision and Recall with respect to the parameter $k$, while $N$ is fixed at $4,000$, are shown in \Cref{app:fig:diversity_k}. We observe that Precision and Recall stabilize as $N$ increases, reaching a plateau around $N=3,000$. Thus, we set the number of samples at $N=4,000$ for subsequent experiments. Additionally, Precision and Recall increase towards $1$ as $k$ increases, aligning with the intuition that the estimated support is more covering as $k$ increases. Therefore, we set $k=4$ as it facilitates a clear differentiation between the two models. We also include the standard deviation for all models described in \Cref{tab:generation_parameters}. Standard deviations are computed by varying the output set, the target set, and both. These results are summarized in \Cref{tab:evaluation_setup}. We recommend consistent references for all models to ensure a fair comparison, as the standard deviation can significantly impact the results.

\begin{table}[b!]
    \centering
    \caption{Standard deviation for seeding different sets of generated samples, different set of references samples. The standard deviation are averaged over the different models and their standard deviation are given. }
    \label{tab:evaluation_setup}
    \begin{tabular}{@{}llll@{}}
        \toprule
                                          & Varying $Q$       & Varying $P$       & Varying $P$ and $Q$ \\ \hline \\[-1em]
        $\sqrt{\Var(\mathrm{Precision})}$ & 0.005 $\pm$ 0.001 & 0.024 $\pm$ 0.010 & 0.019 $\pm$ 0.008   \\
        $\sqrt{\Var{(\mathrm{Recall})}}$  & 0.011 $\pm$ 0.003 & 0.006 $\pm$ 0.002 & 0.013 $\pm$ 0.004   \\\bottomrule
    \end{tabular}
\end{table}

\begin{figure}[H]
    \centering
    \includegraphics[width=0.45\linewidth]{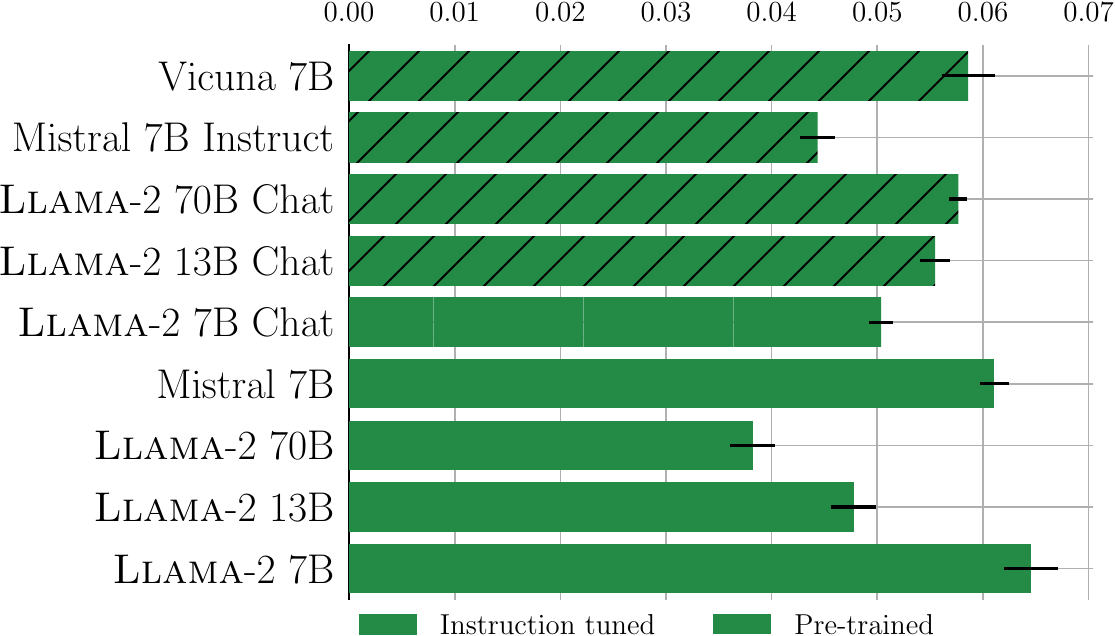}
    \caption{MAUVE score on WebText. Dark bars represents standard error over the generation and reference seeds. No clear distinction between pre-trained and instruction-tuned models.}
    \label{fig:webtext_mauve}
\end{figure}

\paragraph{Additional results.} We report MAUVE scores for WebText on \Cref{fig:webtext_mauve}.
Contrary to what is reflected by Precision and Recall, pre-trained and instruction-tuned models exhibit similar MAUVE scores.

\subsection{Wikipedia Biographies Generation}
\paragraph{Data.} We took inspiration from the WikiBio dataset \citep{wikibio} and extracted Wikipedia articles corresponding to humans. We kept only the articles with a "Good" or "Featured" badge, which are considered as the best quality articles. We then extracted the summaries of these pages to use as reference biographies. Finally, we kept only articles that have at least 80 words and at most 350 words, for a total of 6637 texts. Our processed dataset and the code use for its construction is available in the supplementary materials.

\paragraph{Input formatting.} Pre-trained and chat models are prompted with a small instruction and a variable number of in-context examples. The different prompts are illustrated in \Cref{tab:wiki-input-template}.
\begin{table}[htbp]
    \centering
    \caption{Input Formats for Chat and Non-Chat Models}
    \label{tab:wiki-input-template}
    \begin{tabular}{p{4cm}p{9cm}}
        \toprule
        \textbf{Model}  & \textbf{Wikipedia Biographies input Template}      \\
        \midrule
        \textsc{Llama-2 chat, Mistral chat}
                        & \verb|[INST] Write biographies of various people.| \\
                        & \verb|Here are a few examples:|                    \\
                        & \verb|- Biography of {{name}}: {{content}}|        \\
                        & \verb|{{n times}}|                                 \\
                        & \verb|[/INST]|                                     \\
                        & \verb|- Biography of|                              \\
        \midrule
        \textsc{Vicuna} & \verb|USER: Write biographies of various people.|  \\
                        & \verb|Here are a few examples:|                    \\
                        & \verb|- Biography of {{name}}: {{content}}|        \\
                        & \verb|{{n times}}|                                 \\
                        & \verb|ASSISTANT:|                                  \\
                        & \verb|- Biography of|                              \\
        \midrule
        Non-chat models & \verb|Write biographies of various people.|        \\
                        & \verb|- Biography of {{name}}: {{content}}|        \\
                        & \verb|{{n times}}|                                 \\
        \bottomrule
    \end{tabular}
\end{table}

\paragraph{Generation setup.} We use the default model generation parameters, as described in \Cref{app:webtext}. However, we set the maximum number of tokens generated for all models to 448 for \textsc{Llama-2}-based models (including \textsc{Vicuna}), 432 for \textsc{Mistral} and 390 for \textsc{Pythia}, respecting the average number of tokens per word reported in \Cref{tab:tokens-per-word}.

\subsection{Creative texts generation}

\paragraph{Data.} The full list of creative instructions is presented in Listing 1.

\begin{lstlisting}[caption={List of Creative Prompts},label={app:lst:creative_prompts}]
    Write about a dream you had.
    Create and write about a new character.
    Write about a place you'd love to visit.
    Write about an important life event.
    Write about life 100 years from now.
    Write a story where magic exists in everyday life.
    Write a poem about a personal experience.
    Write a speech for a cause you believe in.
    Write a short mystery story.
    Write a modern day fairy tale.
    Write a story set in a historical period.
    Write a story about a technological advancement.
    Write a letter to your future self.
    Write a story from an animal's perspective.
    Write a week's worth of diary entries for a character.
    Write a short story using mythological characters.
    Write a conversation between two characters.
    Write about a day in your life.
    Write a series of Haikus about seasons.
    Describe a place without naming it.
    Write a news article about an event in your town.
    Write a recipe with a story.
    Write a one-act play.
    Write a story set in a dystopian future.
    Write a satirical essay on a trending topic.
    Write a humorous comic strip.
    Write a piece inspired by nature.
    Write a short story about your favorite fictional character.
    Write instructions for an invented machine.
    Write a story using only metaphors.
    Write a script for a short film.
    Write a conversation in text message format.
    Write a limerick about a funny event.
    Write a story about a journey to space.
    Write a script for a documentary on a subject of your choice.
    Write a eulogy for a character from your favorite book.
    Write a script for a radio show.
    Write a song about a memorable event.
    Write a story with a hidden meaning.
    Write a fable with a moral.
    Write a comedic monologue.
    Write an opinion piece on a current event.
    Write a parody of a popular book or movie.
    Write a story where a character has a magical power.
    Write a story about time travel.
    Write a story based on an object in a mystery box.
    Write a ghost story.
    Describe a haunted house without using cliche descriptions.
    Write a story about a conspiracy theory.
    Write a story about a non-traditional superhero.
\end{lstlisting}

\section{License}

\paragraph{Models weights.} Llama-2 weights are released under the licence available at \url{https://ai.meta.com/llama/license/}. Mistral models and weights are released an Apache 2.0 licence. Vicuna models and weights are released under an Apache 2.0 licence. Pythia models and weights are released under an Apache 2.0 licence.

\paragraph{Data.} The WebText dataset we used is released under a MIT License. Wikipedia's articles are under the reative Commons Attribution-ShareAlike 4.0 International License (CC BY-SA) and the GNU Free Documentation License (GFDL). AG news is available at \url{https://huggingface.co/datasets/ag_news} but no licence are disclosed.
\section{Wikipedia Biographies samples}
\begin{table}
    \centering
    \small % Set font size to small
    \begin{tabularx}{\textwidth}{X}
        Albert Einstein:

        Albert Einstein (born March 14, 1879 - died April 18, 1955) was a German-born theoretical physicist who developed the theory of relativity, one of the two pillars of modern physics (the other being quantum mechanics). His work is also known for its influence on the philosophy of science. He is best known to the general public for his mass-energy equivalence formula E = mc2, which has been dubbed "the world's most famous equation". He received the 1921 Nobel Prize in Physics "for his services to Theoretical Physics, and especially for his discovery of the law of the photoelectric effect", a crucial step in understanding quantum phenomena.

        Einstein was born into a Jewish family in Ulm, Wurttemberg, Germany. In 1896, he enrolled at the Swiss Federal Polytechnic in Zurich to study mathematics and physics. After graduating, he worked as a technical assistant in the patent office in Bern while working on his theories in his free time. He published several papers during this period that brought him to the attention of scientific communities in Europe. In 1905, Einstein published four groundbreaking papers that changed our understanding of space, time, energy, and matter. These papers included special relativity, the photoelectric effect, Brownian motion, and mass-energy equivalence.

        Throughout his career, Einstein held academic positions in Switzerland, Germany, and the United States. He became a professor at the University of Berlin in 1914 but left two years later when Adolf Hitler came to power. He moved to the United States and joined the faculty of the Institute for Advanced Study in Princeton, New Jersey, where he lived until his death. Throughout his life, Einstein continued to publish influential works in both pure and applied physics, including the development of unified field theory, which attempted to explain gravitational forces using electromagnetic principles.

        \\ \hline \\ Marie Curie:

        Marie Sklodowska-Curie (November 7, 1867 - July 4, 1934) was a Polish and naturalized-French physicist and chemist who conducted pioneering research on radioactivity. She was the first woman to win a Nobel Prize, the only person to win twice in multiple sciences, and the only person to win in multiple sciences for a single discovery.
        Born Maria Sklodowska in Warsaw, Poland, she studied mathematics and physics in Paris, where she met her future husband Pierre Curie. Together they discovered the elements polonium and radium, for which they were awarded the Nobel Prize in Physics in 1903. After Pierre's death in 1906, Marie continued her research alone and won the Nobel Prize in Chemistry in 1911 for her discoveries regarding the elements. During the First World War, she established the Radium Institute in Paris to develop mobile radiography units for soldiers. She also worked tirelessly to establish the Curie Institutes in Paris and Warsaw. She continued her scientific work until her death from aplastic anemia, believed to be caused by her long-term exposure to radiation. \\ \hline \\ Marie Curie:

        Marie Sklodowska-Curie (November 7, 1867 - July 4, 1934) was a Polish and naturalized-French physicist and chemist who conducted pioneering research on radioactivity. She was the first woman to win a Nobel Prize, the only person to win two Nobel Prizes in different sciences, and the only person to win a Nobel Prize twice in the same field.
        Born in Warsaw, Poland, Curie moved to Paris at age 24 to study physics and mathematics at the Sorbonne University. She met her future husband, Pierre Curie, while working together in the laboratory. They discovered the elements polonium and radium and shared the 1903 Nobel Prize in Physics with Henri Becquerel for their work on radioactivity. Marie went on to receive the 1911 Nobel Prize in Chemistry for her discovery of the elements radium and polonium and their isolation in pure form.
        In addition to her Nobel Prizes, Curie was also named a Commander of the Legion of Honor and received numerous other honors throughout her career. She became the first professor at the Sorbonne in 1934, just one month before her death from aplastic anemia, believed to be caused by long-term exposure to radiation. Her achievements paved the way for future developments in nuclear science and technology.                                                                                                                                                                                                                                                                                                                                                           \\ \hline \\ Marie Curie:

        Marie Sklodowska-Curie (7 November 1867 - 4 July 1934; French: [maRi sklo'doska]; Polish: [ma'rje skwo'dofska]), born Maria Salomea Sklodowska, was a Polish and naturalized-French physicist and chemist who conducted pioneering research on radioactivity. She is the first woman to win a Nobel Prize, the only person to win two Nobel Prizes in different sciences, and the only person to win in multiple sciences (Physics and Chemistry).
        Born into a Polish family in Warsaw, she studied physics and mathematics at Warsaw University before moving to Paris to continue her studies in science. There, she earned her higher education degrees and conducted groundbreaking research on radioactivity alongside her husband Pierre Curie, whom she married in 1895. Their work led to the discovery of the elements polonium and radium.
        In 1903, they received the Nobel Prize in Physics for their joint research on radiation. Marie continued her work alone following Pierre's death in 1906 and went on to discover the element actinium in 1908. She won the Nobel Prize in Chemistry in 1911 for this achievement. During World War I, she established the Radium Institute in Paris and developed mobile X-ray units for use in treating wounded soldiers on the battlefield. These units became known as Little Curies and saved countless lives.
        Following the war, Marie moved back to Poland and spent time teaching and conducting research at the University of Warsaw. However, due to political instability and financial difficulties, she returned to France in 1934. Tragically, she died there just three months later due to aplastic anemia caused by prolonged exposure to radium. Her achievements and legacy have inspired generations of scientists and continue to shape\end{tabularx}
    \caption{Generated biographies from Mistral Instruct v0.2, with 1 in-context example. The model generates always the same biography, illustrated by a low Recall. The precision is also very low, because Marie-Curie and Einstein are scientific personalities, who are not widely represented in the reference dataset, compared to other occupations.}
\end{table}
\begin{table}
    \centering
    \small % Set font size to small
    \begin{tabularx}{\textwidth}{X}
        Herman Wirth:

        Hermann Wirth (1 September 1885 - 12 March 1981) was a German historian and philologist. He was professor of Indo-European studies at Bonn University from 1920 to 1933, where he taught among others Hans Freudenthal, Ernst Haefliger, Erich Schenk, Wilhelm Schulze, Johannes Schmidt, Werner Voigt, Heinrich Vietzen, Wolfgang Golther, Martin Doutte, Fritz Graf, Paul Maas, Karl Meisenheim, Rudolf Much, Otto Neugebauer, Bruno Obermaier, Helmut Paech, Richard Pittioni, Gerhard Pommeranz-Wardbor, Ulrich Raulff, Albert Siebeck, Kurt Steinmann, Alfred Thayer Mahan, Max Vasmer, Eduard Frauwallner, Robert Havemann, and Hermann Schneider. From 1933 to 1935 he directed the Kaiser Friedrich Museum in Berlin. During World War II he continued his research in Paris under the pseudonym "Dr. Schmitt". He was the founder of the European School of Folklore Studies (Europaische Hochschule fur Volkskunde) in Salzburg.
        In 1928 Wirth founded the journal Zeitschrift fur vergleichende Sprachforschung und verwandte Gebiete ("Journal for Comparative Philology"). Its editorial board included prominent linguists such as Georges Dumezil, Leo Wiener, Karl Brugmann, Johann Christian Bekker, Emil Siegfried Engelsvater, Alexander Kappelmacher, Carl Darling, Johannes Droys, Jacob Grimm, Hugo Koch, Georg Curtius, Julius Pokorny, Gustav Reichen, and Bernhard Wiese.

        \#\#\# Other languages

        * Arabic:                                           \\ \hline \\ George Cunningham:

        George John Cunningham (23 February 1905 - 22 December 1995) was a Scottish footballer who spent all but two seasons of his senior football career playing for Manchester City. He scored 12 goals from 32 appearances in representative matches for Scotland, and was inducted into the English Football Hall of Fame in 2002. He went on to manage clubs such as Burnley and Derby County, and was part of the management team when the latter won promotion to the Premier League in 2007.

        \#\#\# Sports                                       \\ \hline \\ Yuriy Kuleshov:

        Yury Petrovich Kuleshov (Ukrainian: Iurii Petrovich Kuleshov; 2 January 1922 - 17 February 2009) was a Ukrainian football defender. He was a squad member for the Soviet Union national football team at the 1952 Summer Olympics. He was a former Dynamo Kyiv - 1942, FC Dynamo 1943, 1942, 1944, Dynamo Kyiv 1945.
        He was awarded the title Master of Sport USSR 1956. \\ \hline \\ Danielle Bregoli:

        Danielle Michelle Bregoli (/'breg@li:/, BRUG-@-lee; born March 26, 2003), better known online as Bhad Bhabie /baed b@bi:/, is an American rapper, singer, songwriter, internet personality and businesswoman. Born in Boynton Beach, Florida, Bregoli gained fame through her appearance on Dr Phil in September 2016, wherein she told the audience that she wanted to "clap back" at people who disrespected her mother. This phrase subsequently became popularized and has been referenced in numerous media outlets, songs, TV shows and films since then. Since appearing on Dr Phi, Bregoli has achieved commercial success as a recording artist, having reached number one on the Billboard Hot 100 chart with "These Heaux". As of May 2022, Bregoli has amassed over 1 billion views across her YouTube channel.
        In addition to music, Bregoli has launched clothing lines and beauty brands, as well as acting roles.\end{tabularx}
    \caption{Generated biographies from Llama-2-7b, with 11 in-context examples. The samples are of low quality, illustrated by a low Precision. On the contrary, the Recall is high, as the model generates a wide variety of biographies.}
\end{table}
\end{document}